\documentclass[10pt,twocolumn,letterpaper]{article}

\usepackage[pagenumbers]{cvpr} %

\usepackage{graphicx}
\usepackage{amsmath}
\usepackage{amssymb}
\usepackage{booktabs}
\usepackage{float}
\usepackage{enumitem}

\usepackage[pagebackref,breaklinks,colorlinks]{hyperref}

\usepackage[capitalize]{cleveref}
\crefname{section}{Sec.}{Secs.}
\Crefname{section}{Section}{Sections}
\Crefname{table}{Table}{Tables}
\crefname{table}{Tab.}{Tabs.}

\def\cvprPaperID{4797} %
\def\confName{CVPR}
\def\confYear{2023}

\usepackage{color}
\usepackage{soul}
\usepackage{multirow}
\usepackage{xcolor}

\newlength\savewidth\newcommand\shline{\noalign{\global\savewidth\arrayrulewidth\global\arrayrulewidth 1pt}\hline\noalign{\global\arrayrulewidth\savewidth}}

\definecolor{turquoise}{cmyk}{0.65,0,0.1,0.1}
\definecolor{purple}{rgb}{0.65,0,0.65}
\definecolor{darkgreen}{rgb}{0.0, 0.5, 0.0}
\definecolor{darkred}{rgb}{0.5, 0.0, 0.0}
\definecolor{darkblue}{rgb}{0.0, 0.0, 0.5}
\definecolor{blue}{rgb}{0.0, 0.0, 1.0}
\definecolor{orange}{rgb}{1.0,0.5,0.0}

\newcommand{\hide}[1]{{}}

\makeatletter
\renewcommand{\paragraph}{%
  \@startsection{paragraph}{4}%
  {\z@}{2.5ex \@plus 1ex \@minus .2ex}{-1em}%
  {\normalfont\normalsize\bfseries}%
}
\makeatother

\newif\ifproofread

\newcommand\PHLab{{PixHt-Lab}} %
\newcommand\SN{{SSG++}}        %

\begin{document}

\title{\PHLab: Pixel Height Based Light Effect Generation for Image Compositing}

\author{Yichen Sheng\\
Purdue University\\
\and
Jianming Zhang\\
Adobe Inc. \\
\and 
Julien Philip \\
Adobe Inc. \\
\and 
Yannick Hold-Geoffroy \\
Adobe Inc. \\
\and 
Xin Sun \\
Adobe Inc. \\
\and 
He Zhang \\
Adobe Inc. \\
\and 
Lu Ling \\
Purdue University \\
\and 
Bedrich Benes \\
Purdue University \\
}

\twocolumn[{%
\renewcommand\twocolumn[1][]{#1}%
\maketitle
\vspace{-5mm}
\setlength{\tabcolsep}{0pt}
\centering
\begin{tabular}{cccc}
    \includegraphics[width=0.25\textwidth, trim={2.0cm 1.0cm 2.0cm 3.0cm},clip]{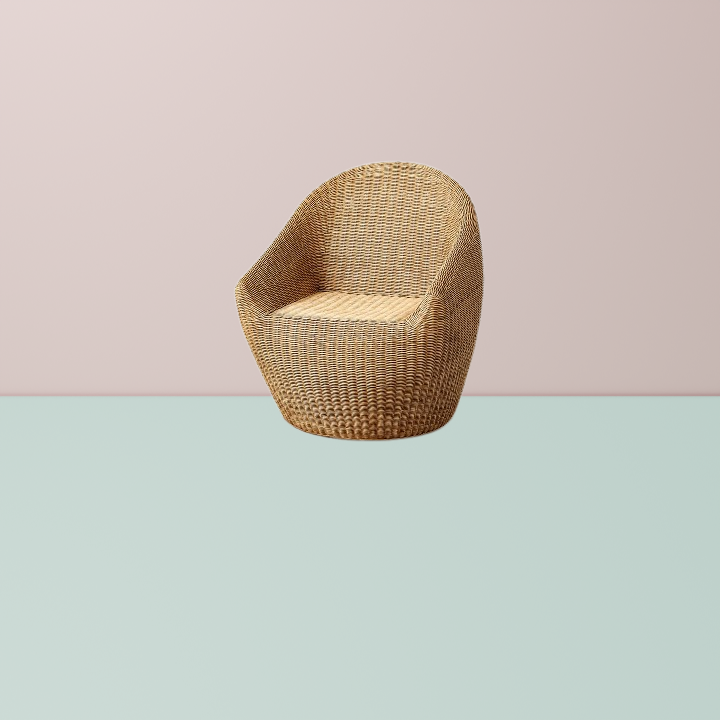} &
    \includegraphics[width=0.25\textwidth, trim={2.0cm 1.0cm 2.0cm 3.0cm},clip]{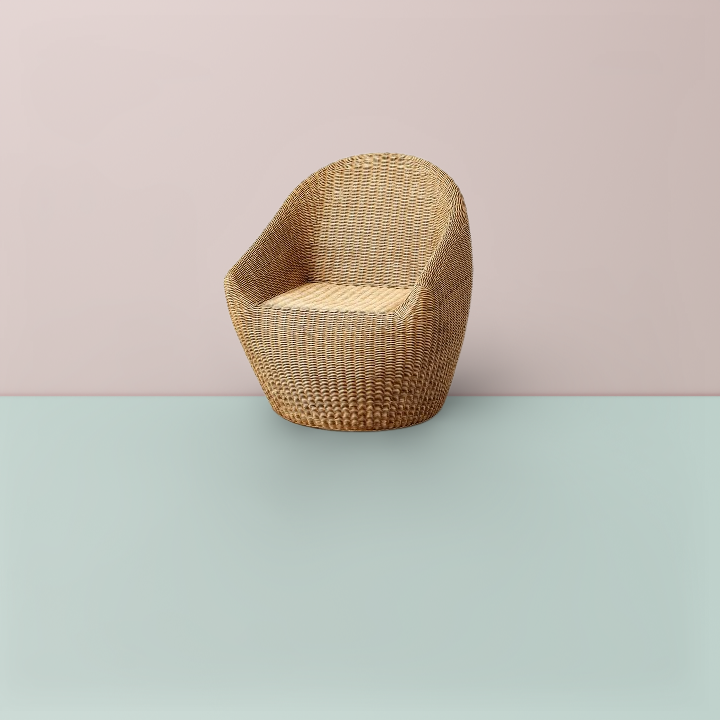} &
    \includegraphics[width=0.25\textwidth, trim={2.0cm 1.0cm 2.0cm 3.0cm},clip]{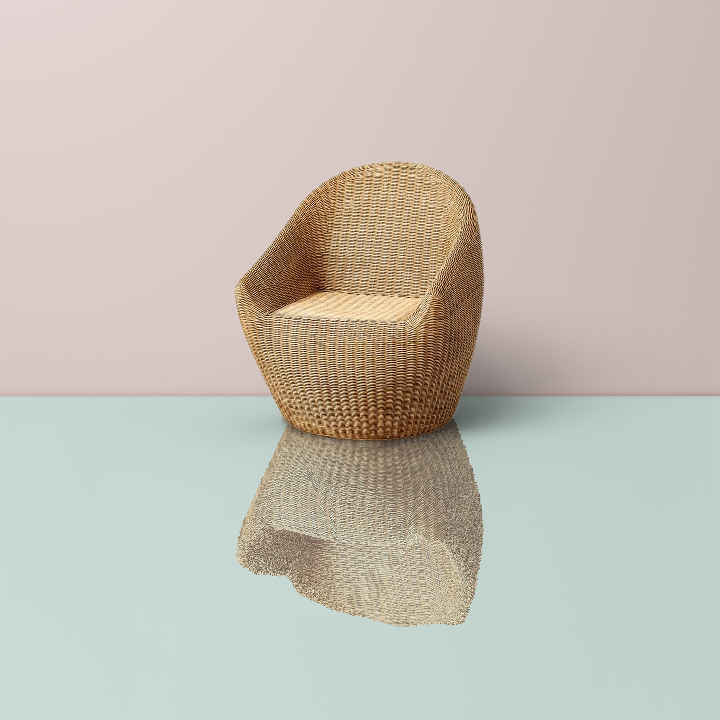} &
    \includegraphics[width=0.25\textwidth, trim={2.0cm 1.0cm 2.0cm 3.0cm},clip]{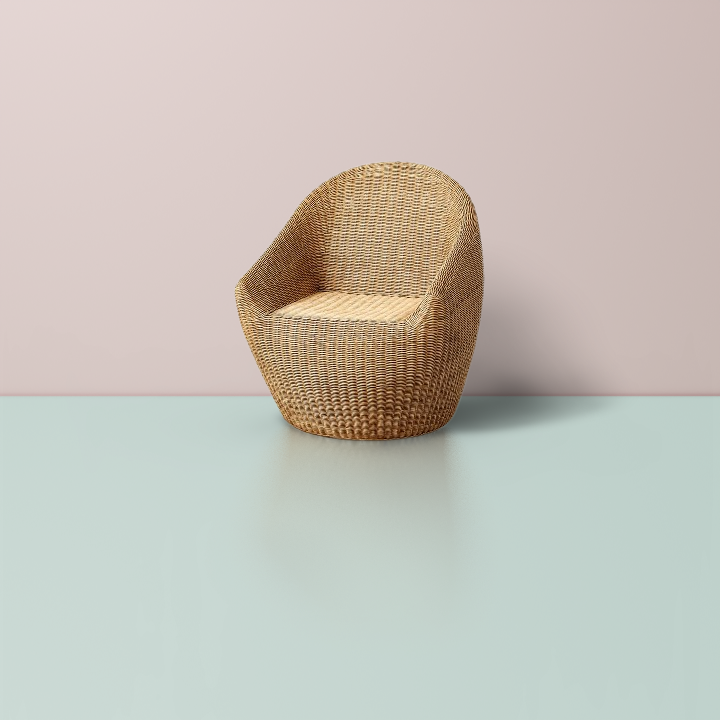} \\
    (a) cutout & (b) SSN~\cite{sheng2021ssn} & (c) SSG~\cite{sheng2022controllable} & (d) \PHLab{} (ours)
    
\end{tabular}%

\captionof{figure}{
\textbf{\PHLab} renders realistic reflection and soft shadows on general shadow receivers for 2D cutouts. (a) shows the object cutout and composition background. (b) SSN~\cite{sheng2021ssn} cannot render soft shadows on walls due to its ground plane assumption. 
(c) SSG~\cite{sheng2022controllable} renders specular reflection, but the shadow on the wall is uniformly softened. (d) Our \PHLab{} renders realistic soft shadows on the wall guided by 3D-aware buffer channels. The shadow is softened according to the background geometry with more realistic details. \PHLab{} also renders realistic reflections with physically-based surface materials.} \label{fig:teaser}

\bigbreak
}]

\begin{abstract}
Lighting effects such as shadows or reflections are key in making synthetic images realistic and visually appealing. To generate such effects, traditional computer graphics uses a physically-based renderer along with 3D geometry. To compensate for the lack of geometry in 2D Image compositing, recent deep learning-based approaches introduced a pixel height representation to generate soft shadows and reflections. However, the lack of geometry limits the quality of the generated soft shadows and constrain reflections to pure specular ones.
We introduce \PHLab{}, a system leveraging an explicit mapping from pixel height representation to 3D space. Using this mapping, \PHLab{} reconstructs both the cutout and background geometry and renders realistic, diverse, lighting effects for image compositing. Given a surface with physically-based materials, we can render reflections with varying glossiness. To generate more realistic soft shadows, we further propose to use 3D-aware buffer channels to guide a neural renderer. Both quantitative and qualitative evaluations demonstrate that \PHLab{} significantly improves soft shadow generation. 
\end{abstract}

\section{Introduction}
Image compositing is a powerful tool widely used for image content creation, combining interesting elements from different sources to create a new image. 
One challenging task is adding lighting effects to make the compositing realistic and visually appealing. 
Lighting effects often involve complex interactions between the objects in the compositing, so their manual creation is tedious. It requires a significant amount of effort, especially for soft shadows cast by area lights and realistic reflections on the glossy surface with the Fresnel effect\cite{frensnel}. 

Many methods that generate lighting effects for 3D scenes have been well-studied~\cite{kajiya1986rendering}, but 3D shapes are often unavailable during image compositing. 
Recent advancements in deep learning made significant progress in lighting effect generation in 2D images, especially for shadow generation. A series of generative adversarial networks (GANs) based methods~\cite{hu2019mask, liu2020arshadowgan, zhang2019shadowgan,wang2020people} have been proposed to automatically generate hard shadows to match the background by training with pairs of shadow-free and shadow images.
Those methods only focus on hard shadow generation, and their generated hard shadow cannot be edited freely. 
More importantly, the light control of those methods is implicitly represented in the background image. 
In real-world image creation scenarios, however, the background is often well-lit or even in pure color under a studio lighting setting, making those methods unusable. 
Also, editing the shadow is often needed on a separate image layer when the image editing is still incomplete.

To address these issues, a recent work SSN~\cite{sheng2021ssn} proposes to learn the mapping between the image cutouts and the corresponding soft shadows based on a controllable light map. 
Although it achieves promising results, it assumes that the shadow receiver is just a ground plane and the object is always standing on the ground, which limits its practical usage.
This limitation is addressed by SSG~\cite{sheng2022controllable}, which proposes a new 2.5D representation called pixel height, which is shown to be better suited for shadow generation than previous 2.5D presentations like depth map.
Hard shadow on general shadow receivers can be computed by a ray tracing algorithm in the pixel-height space. A neural network renderer is further proposed to render the soft shadow based on the hard shadow mask. 
It achieves more controllability and it works in more general scenarios, but the lack of 3D geometry guidance makes the soft shadows unrealistic and prone to visual artifacts when they are cast on general shadow receivers like walls. 
In addition, SSG proposes an algorithm to render the specular reflection by flipping the pixels according to their pixel height. However, the use case is very limited as it cannot be directly applied to simulate realistic reflection effects on more general materials (see Fig.~\ref{fig:teaser} (c)).   

We introduce a controllable pixel height based system called \PHLab{} that provides lighting effects such as soft shadows and reflections for physically based surface materials. 
We introduce a formulation to map the 2.5D pixel height representation to the 3D space. 
Based on this mapping, geometry of both the foreground cutout and the background surface can be directly reconstructed by their corresponding pixel height maps. 
As the 3D geometry can be reconstructed, the surface normal can also be computed. 
Using a camera with preset extrinsic and intrinsics, light effects, including reflections, soft shadows, refractions, etc., can be rendered using classical rendering methods based on the reconstructed 3D geometry or directly in the pixel height space utilizing the efficient data structure (See Sec.~\ref{sec:3D_Render}) derived from the pixel height representation.

As the soft shadow integration in classical rendering algorithms is slow, especially for large area lights, we propose to train a neural network renderer \SN{} guided by 3D-aware buffer channels to render the soft shadow on general shadow receivers in real time. Quantitative and qualitative experiments have been conducted to show that the proposed \SN{} guided by 3D-aware buffer channels significantly improve the soft shadow quality on general shadow receivers than previous soft shadow generation works. 

Our main contributions are:
\begin{itemize}[noitemsep]
  \item A mapping formulation between pixel height and the 3D space. Rendering-related 3D geometry properties, e.g., normal or depth, can be computed directly from the pixel height representation for diverse 3D effects rendering, including reflection and refraction. 
  \item A novel soft shadow neural renderer, \SN{}, guided by 3D-aware buffer channels to generate high-quality soft shadows for general shadow receivers in image composition.
\end{itemize}

\section{Previous Work}
\paragraph{Single Image 3D Reconstruction} 
Rendering engines can be used to perform image composition. 
However, they require a 3D reconstruction of the image, which is a challenging problem.
Deep learning-based methods \cite{sinha2017surfnet, kong2017using, pontes2017compact, pontes2018image2mesh} have been proposed to perform dense 3D reconstruction via low dimensional parameterization of the 3D models, though rendering quality is impacted by the missing high-frequency features. 
Many single-image digital human reconstruction methods \cite{bogo2016keep, lassner2017unite, kanazawa2018end, zhang_real-time_2017, xiang_monocular_2019, pavlakos_expressive_2019, xu_denserac_2019, saito2019pifu, zheng2019deephuman, saito2020pifuhd} show promising results, albeit they assume specific camera parameters. Directly rendering shadows on their reconstructed 3D models yields hard-to-fix artifacts \cite{sheng2022controllable} in the contact regions, between the inserted object and the ground. 
More importantly, those methods cannot be applied to general objects, which limits their use for generic image composition. 

\paragraph{Single Image Neural Rendering} Image harmonization blends a cutout within a background in a plausible way. Classical methods achieve this goal by adjusting the appearance statistics \cite{pitie_n-dimensional_2005, reinhard_color_2001, jia_drag-and-drop_2006, perez2003poisson, tao_error-tolerant_2010}.
Recently, learning-based methods~\cite{tsai_deep_2017, jiang_ssh_2021, cong2020dovenet, sofiiuk2021foreground, jiang2021ssh,ling2021region,valanarasu2022interactive} trained with augmented real-world images were shown to provide more robust results. 
However, these methods focus on global color adjustment without considering shadows during composition. 
Single image portrait relighting methods~\cite{zhou_deep_2019, shu_portrait_2017, sun_single_2019} adjust the lighting conditions given a user-provided lighting environment, although they only work for human portraits. 
\cite{griffiths2022outcast} considers the problem of outdoor scene relighting from a single view using intermediary predicted shadow layers, which could be trained on cutout objects, but their method only produces hard shadows. 
Neural Radiance Field based methods (e.g., \cite{srinivasan_nerv_2021, mildenhall_nerf_2022, martin_brualla_nerf_2021, avidan_nerf_2022}) propose to implicitly encode the scene geometry, but require multiple images as input.

\paragraph{Soft Shadow Rendering} is a well-studied technique in computer graphics, whether for real-time applications \cite{assarsson_geometry-based_nodate, donnelly_variance_2006, schwarz_bitmask_2007, williams_casting_nodate, annen_real-time_2008, chan_rendering_nodate, fernando_percentage-closer_2005, guennebaud_real-time_nodate, guennebaud_high-quality_2007, soler_fast_1998, ng_all-frequency_nodate, sen2003shadow, SD02, reeves1987rendering} or global illumination methods \cite{cook_computer_1984, kajiya_rendering_1986, sillion_global_1991, westin_predicting_nodate}. It requires exact 3D geometry as input, preventing its use for image composition. 

\begin{figure*}[t]
    \centering
    \includegraphics[width=0.99\linewidth]{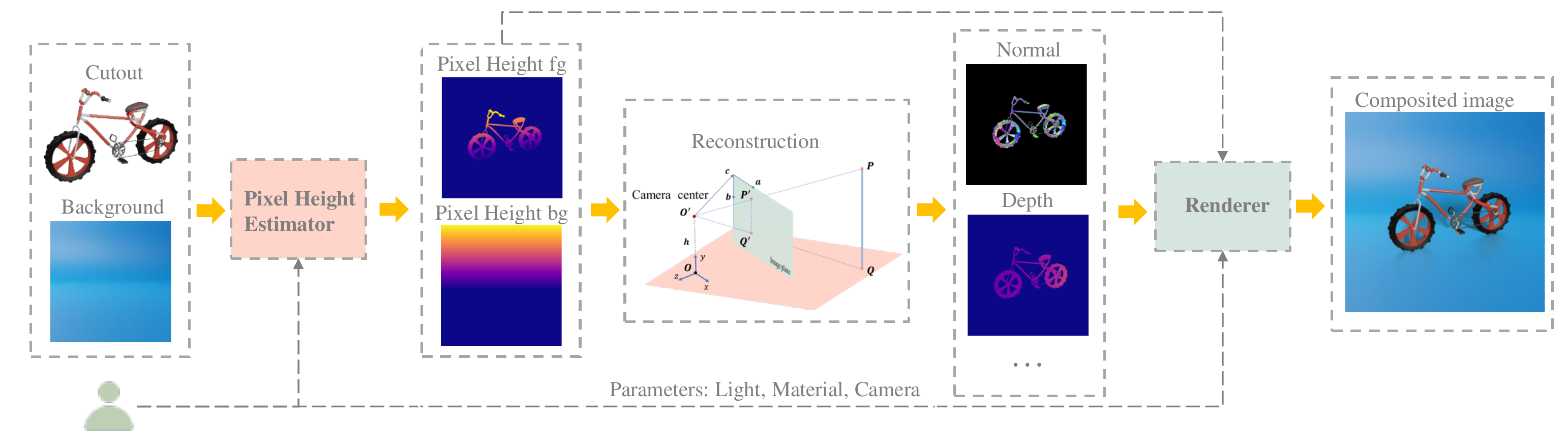}
    \caption{\textbf{System overview of \PHLab}. Given a 2D cutout and background, the pixel height maps for the cutout and background can be either predicted by a neural network~\cite{sheng2022controllable} or labeled manually by the user. 3D scene information and the relevant buffer channels can then be computed from pixel height based on our formulation presented in Sec.~\ref{sec:PH_to_3D}. Finally, our neural renderer \SN{} renders the requested lighting effects using the buffer channels (see Sec.~\ref{sec:3D_Render}).}
    \label{fig:pipeline}
\end{figure*}

Recent neural rendering methods can address the limited input problem and render shadows for different scenarios. 
Scene level methods~\cite{philip2021free, philip2019multi} show promising results but require multiple input images.
Generative adversarial networks (GANs) have achieved significant improvements on image translation tasks~\cite{isola2017image, liu2018auto}, and subject-level shadow rendering methods~\cite{hu2019mask, liu2020arshadowgan, zhang2019shadowgan,wang2020people} propose to render shadow using GANs.
Unfortunately, these methods have two main limitations: they can only generate hard shadows, and prevents user editability, which is desired for artistic purposes. 
SSN \cite{sheng2021ssn} proposed a controllable soft shadow generation method for image composition, but is limited to the ground plane and cannot project shadows on complex geometries. 
Recently, SSG~\cite{sheng2022controllable} further proposed a new representation called \textbf{pixel height} to cast soft shadows on more general shadow receivers and render specular reflection on the ground. 
Unfortunately, the shadow quality of SSG degrades on complex geometries as they are not explicitly taken into account by the network. 
Furthermore, its reflection rendering is limited to specular surfaces. 
In contrast, our proposed \SN{} is guided by 3D geometry-aware buffer channels that can render more realistic soft shadows on generic shadow receivers. 
We further connect the pixel height representation to 3D by using an estimated per-pixel depth and normal, increasing the reflections' realism. 

\section{Method}
We propose a novel algorithm (Fig.~\ref{fig:pipeline}) to render reflection and soft shadow to increase image composition realism based on pixel height~\cite{sheng2022controllable}, which has been shown to be more suitable for shadow rendering. Pixel height explicitly captures the object-ground relation, and thus it better keeps the object uprightness and the contact point for shadow rendering. Moreover, it allows intuitive user control to annotate or correct the 3D shape of an object.

Our first key insight is that 3D information that highly correlates with rendering many 3D effects, e.g., spatial 3D position, depth, normal, etc., can be recovered by a mapping (see Sec.~\ref{sec:PH_to_3D}) given the pixel height representation. 
The second key idea is that soft shadows on general shadow receivers are correlated with the relative 3D geometry distance between the occluder and the shadow receiver. 
Based on the mapping from pixel height to 3D, several geometry-aware buffer channels (see Sec.~\ref{sec:Buffers}) are proposed to guide the neural renderer to render realistic soft shadows on general shadow receivers. Moreover, acquiring the 3D information enables rendering additional 3D effects, e.g., reflections and refractions.

Fig.~\ref{fig:pipeline} shows the overview of our method. Given the 2D cutout and background, the pixel height maps for the cutout and background can be either predicted by a neural network~\cite{sheng2022controllable} or labeled manually by the user. 3D geometry that is used in rendering can be computed using our presented method (see Sec.~\ref{sec:PH_to_3D}). 
Finally, our renderer (see Sec.~\ref{sec:3D_Render}) can add 3D effects to make the image composition more realistic.

\subsection{Connecting 2.5D Pixel Height to 3D} \label{sec:PH_to_3D}

\begin{figure}[t]
    \centering
    \includegraphics[width=.9\linewidth]{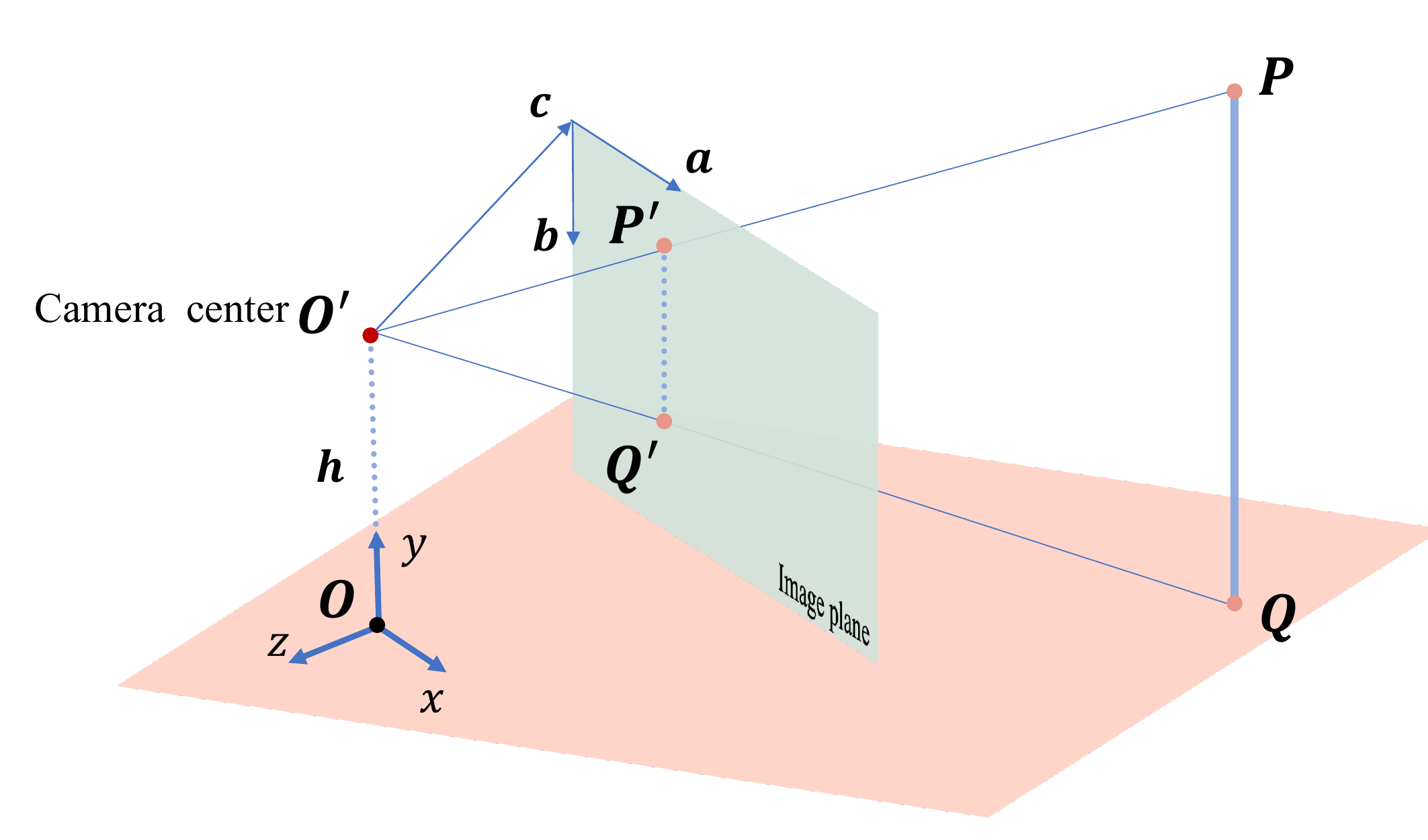}
    \caption{\textbf{Connecting pixel height to 3D.} Given the camera at the center $O'$ and its foot point $O$, a point $P$ in 3D space with its foot point $Q$. $P'$ and $Q'$ are their projection positions on the image plane. $c$ is the vector from $O'$ to the up left corner of the image plane. $a$ and $b$ are the right directions and down direction vector relative to the image plane.}
    \label{fig:3d_reconstruct}
\end{figure}

Here we describe the equation that connects 2.5D pixel height to its corresponding 3D point, and Fig.~\ref{fig:3d_reconstruct} shows the camera and relevant geometry and variables. 
We define $O$, the foot point of $O'$, as the origin of the coordinate system. For convenience, we define the camera intrinsics by three vectors: 1)~the vector $c$ from the camera center~$O'$ to the top left corner of the image plane, 2)~the right vector of the image plane~$a$, and 3)~the down vector of the image plane~$b$. Any point $P$ and its foot point $Q$ can be projected by the camera centered at $O'$. The points $P$ and $Q$ projected on the image plane are denoted as $P'$ and $Q'$. 
\begin{equation} \label{eq: proj}
    \begin{bmatrix}
    x_{O'} \\
    y_{O'} \\
    z_{O'}
    \end{bmatrix} + 
    \begin{bmatrix}
    x_a & x_b & x_c \\
    y_a & y_b & y_c \\
    z_a & z_b & z_c \\
    \end{bmatrix}
    \begin{bmatrix}
    u_{P'} \\
    v_{P'} \\ 
    1
    \end{bmatrix} w
    = 
    \begin{bmatrix}
    x_P \\
    y_P \\
    z_P
    \end{bmatrix}
\end{equation} 

\begin{equation} \label{eq: solve_foot}
y_{o'} + 
\begin{bmatrix}
y_a & y_b & y_c  
\end{bmatrix} 
\begin{bmatrix}
u_{Q'} \\ v_{Q'} \\ 1  
\end{bmatrix} 
w = 
y_Q
\end{equation}

The relationship between a 3D point $P$ and its projection $P'$ is described by the projection Eq.~\ref{eq: proj} under the pinhole camera assumption. From the definition of pixel height representation, the foot point $Q'$ of $P'$ in the image space is on the ground plane, i.e., $y_{Q} = 0$. Solving Eq.~\ref{eq: solve_foot} provides $w$: 
\begin{equation}\label{eq: w}
    w = \frac{-y_{o'}}{y_a u_{q'} + b_y v_{q'} + y_c} 
\end{equation} 
The pixel height representation has no pitch angle assumption, thus $P$ can be directly computed using the $w$ in Eq.~\ref{eq: w}. By re-projecting the $P'$ back, the 3D point $P$ can be calculated as
\begin{equation}\label{eq: P}
\footnotesize
P = 
    \begin{bmatrix}
    x_{O'} \\
    y_{O'} \\
    z_{O'}
    \end{bmatrix}
    + 
    \begin{bmatrix}
    x_a & x_b & x_c \\
    y_a & y_b & y_c \\
    z_a & z_b & z_c \\
    \end{bmatrix}
    \begin{bmatrix}
    u_{P'} \\
    v_{P'} \\ 
    1
    \end{bmatrix} 
    \frac{-h}{y_a u_{Q'} + y_b v_{Q'} + y_c}.
\end{equation}

\begin{figure}[t]
\small
\setlength{\tabcolsep}{5pt}
\centering
\begin{tabular}{cc}
    \captionsetup{type=figure}
    \includegraphics[width=0.49\linewidth]{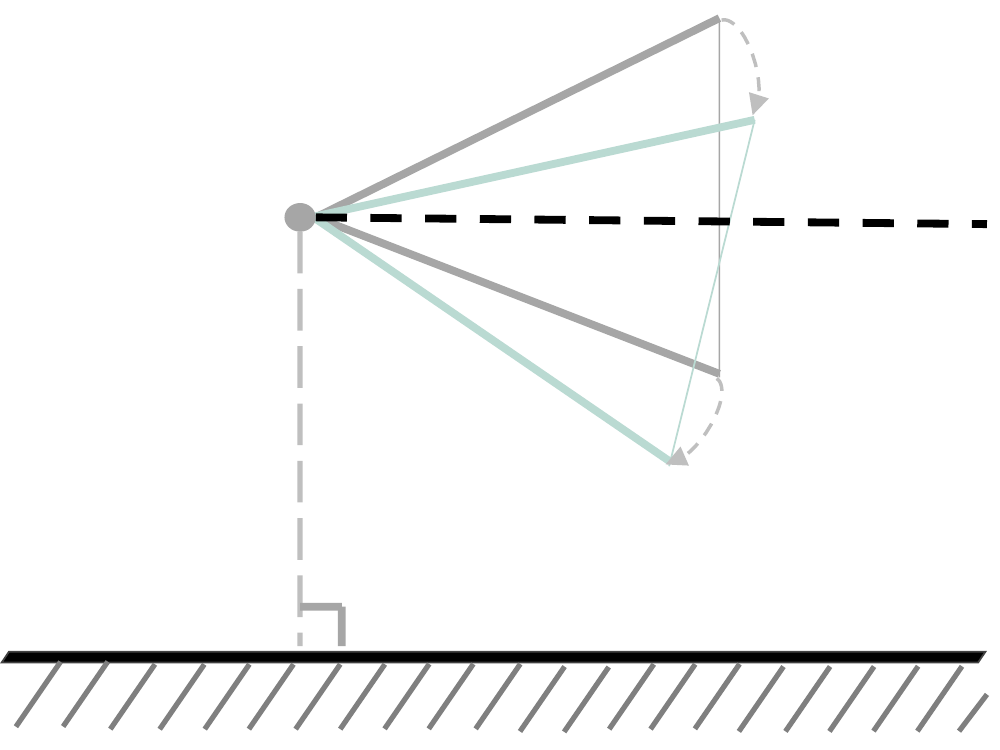} & \includegraphics[width=0.33\linewidth]{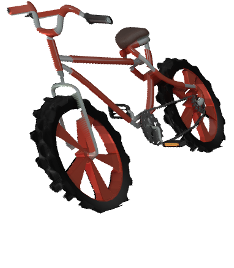} \\
    (a) Classical camera & (b) Reconstructed by (a) \\
    \includegraphics[width=0.49\linewidth]{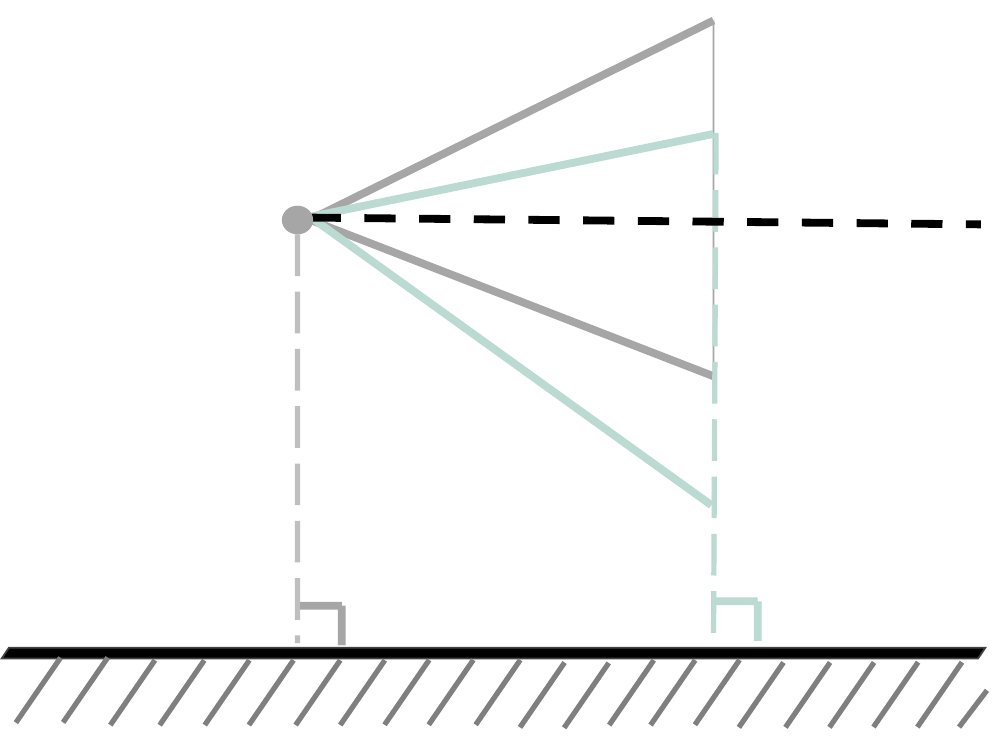} & 
    \includegraphics[width=0.33\linewidth]{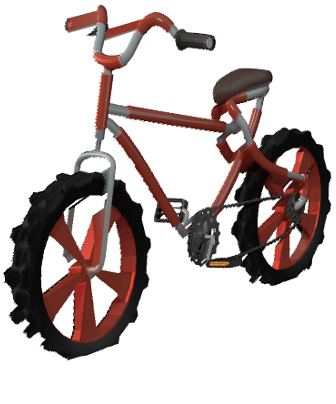}\\
    (c) Tilt-shift camera & (d) Reconstructed by (c) 
\end{tabular}
\captionof{figure}{\textbf{Tilt shift camera model}. Horizon position is a controllable parameter to change the shadow perspective shape. (a) shows changing the horizon is equivalent to changing the pitch in the classical model. (b) shows the reconstructed 3D model. (c) shows we use a tilt-shift camera. (d) shows the 3D vertical line can be preserved to be perpendicular to the ground after reconstruction.}\label{fig:axis_align_camera}
\end{figure}

Horizon controllability is important as different horizons will change the soft shadow distortion. However, as shown in Fig.~\ref{fig:axis_align_camera}, changing the horizon will change the camera pitch, which violates the no pitch assumption from pixel height representation~\cite{sheng2022controllable} and leads to tilted geometry.
To resolve the issue, we propose to use a tilt-shift camera model for our application. 
When the user changes the horizon, the vector $c$ in Fig.~\ref{fig:3d_reconstruct} will move vertically to align the horizon to keep the image plane perpendicular to the ground. In this way, the no-pitch assumption is preserved for the correct reconstruction, as shown in Fig.~\ref{fig:axis_align_camera}~(d). 

\subsection{3D Buffer Channels for Soft Shadow Rendering} \label{sec:Buffers}

Our methods can be applied to arbitrary shape lights, but for discussion purposes, we assume the light for our discussion is disk shape area light. It is challenging to render high-quality soft shadows for general shadow receivers given only image cutouts and the hard shadow mask, as the soft shadow is jointly affected by multiple factors: the light source geometry, the occluder, the shadow receiver, the spatial relationship between the occluder and the shadow receiver, etc. SSG~\cite{sheng2022controllable} is guided by the cutout mask, the hard shadow mask, and the disk light radius as inputs. The shadow boundary is softened uniformly (see Figs.~\ref{fig:teaser} and~\ref{fig:ablation}) as SSG is unaware of the geometry-related information relevant to soft shadow rendering. We propose to train a neural network \SN{} to learn how these complex factors will jointly affect the soft shadow results. 

\begin{figure}
    \centering
    \includegraphics[width=\linewidth, trim={0 2.0cm 0 0},clip]{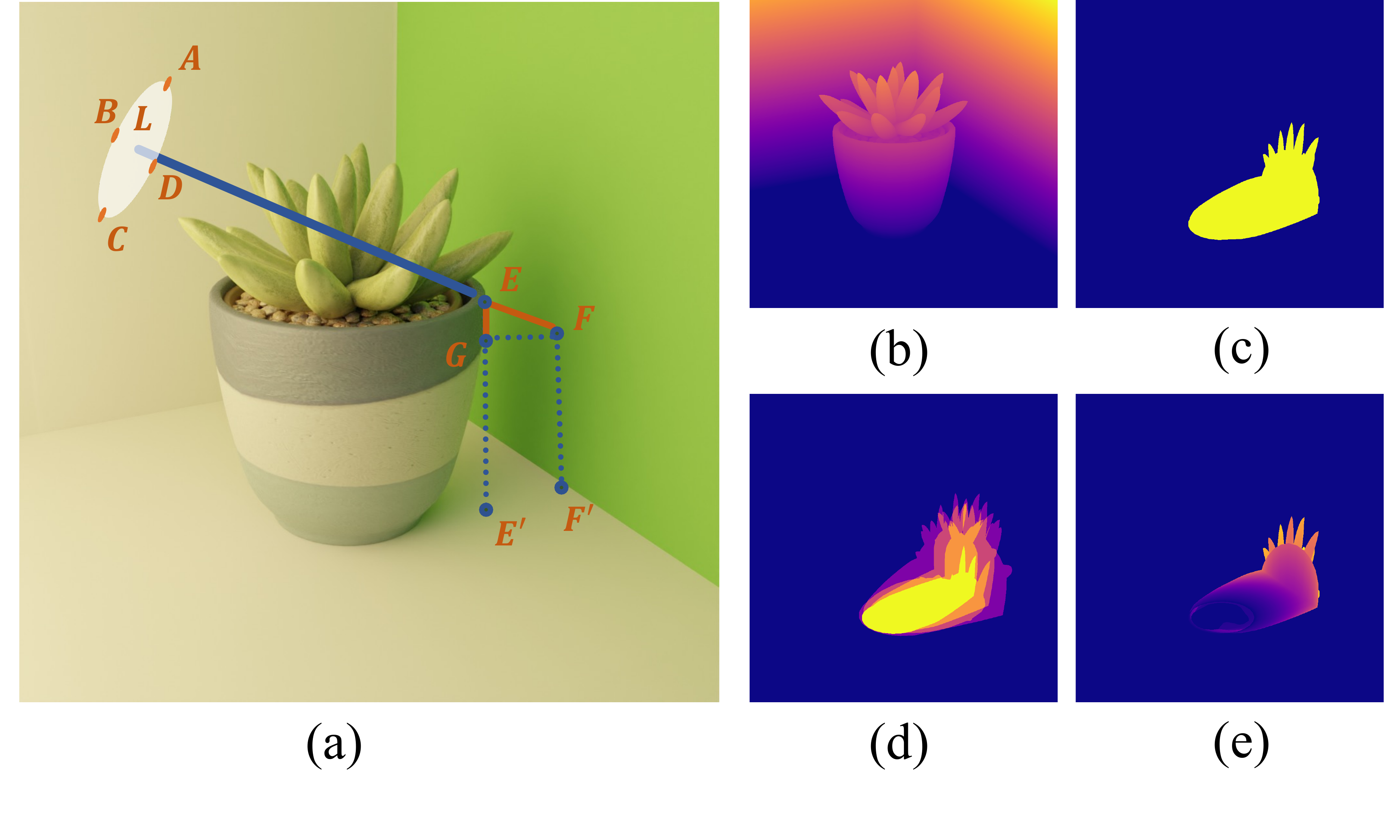}  %
    \caption{\textbf{Buffer channels}. (a) illustrates how the buffer channels are computed. See the text for more details. (b) shows the pixel height maps of the foreground cutout and the background. (c) is the hard shadow cast by the center of the disk area light $L$. (d) is the sparse hard shadows map cast by $A, B, C, D$, which are four extreme points of the area light $L$. (e) is the distance between $EF$ in pixel height space.}
    \label{fig:buffer channels}
\end{figure}

\paragraph{3D-Aware Buffer channels.} Our \SN{} takes several 3D-aware buffer channels (see Fig.~\ref{fig:buffer channels}) relevant to soft shadow rendering. The buffer channels are composed of several maps: the cutout pixel height map; the gradient of the background pixel height map; the hard shadow from the center of the light $L$; the sparse hard shadows map;
the relative distance map between the occluder and the shadow receiver in pixel height space. For illustration purposes, we composite foreground pixel height and background pixel height in Fig.~\ref{fig:buffer channels} (b).

The cutout pixel height and background pixel height map describe the geometry of the cutout and the background.
In our implementation, we use the gradient map of the background pixel height as input to make it translation invariant. The pixel height gradient map will capture the surface orientation similar to a normal map.

The sparse hard shadows map can also guide the network to be aware of the shadow receiver's geometry. Another important property of this channel is that the sparse hard shadows describe the outer boundary of the soft shadow. The overlapping areas of the sparse hard shadows are also a hint of darker areas in the soft shadow. Experiments in Sec.~\ref{Sec:Exp} show this channel plays the most important role among all the buffer channels. The four sparse hard shadows are sampled from the four extreme points of the disk light $L$ as shown in Fig.~\ref{fig:buffer channels}. 

The relative distance map in pixel height space defines the relative spatial distance between the occluder and the shadow receiver.
The longer the distance, the softer the shadow will be in general. 
This channel guides the network to pay attention to shadow regions that have high contrast. 
The formal definition of the relative distance in pixel height space is:
$||(u_p, v_p, h_p)-(u_q, v_q, h_q)||_2^2$, where $p, q$ are two points, $u, v$ are the coordinates in the pixel space, $h$ is the pixel height. 

\paragraph{Dataset and Training.}
We follow SSN~\cite{sheng2021ssn} and SSG~\cite{sheng2022controllable} to generate a synthetic dataset to train \SN{}. In practice, we randomly picked 100 general foreground objects from ModelNet~\cite{wu20153d} and ShapeNet~\cite{chang2015shapenet} with different categories, including human, plants, cars, desks, chairs, airplanes, etc. 
We also picked different backgrounds: ground plane, T shape wall, Cornell box, and curved plane. To cover diverse relative positions between the foreground and background, we randomly generate 200 scenes from the selected foreground and background objects.
For each scene, we further randomly sampled 100 lights per scene from a different direction with random area light sizes. In total, the synthetic dataset has 20k training data.  
\SN{}{} follows the SSG architecture.
We implement the \SN{} using PyTorch\cite{paszke2019pytorch}. The training takes 8 hrs on average to converge with batch size 50 and learning rate $2e^{-5}$ in a RTX-3090.

\subsection{Ray Tracing in Pixel Height Representation} \label{sec:3D_Render}
Eq.~\ref{eq: proj} in Sec.~\ref{sec:PH_to_3D} connects the pixel height to 3D. Although we do not know the camera extrinsic and intrinsic for the image cutout or the background, we can use a default camera to reconstruct the scene, given the pixel height.   
When the 3D position for the 2D pixel can be computed, the surface normal can be approximated if we assume neighborhood pixels are connected. 
When the surface normal can be reconstructed, 3D effects, including reflection, refraction and relighting, can be rendered if surface materials are given.  

One can perform the 3D effects rendering using a classical graphics renderer to render lighting effects for image compositing. We noticed that the pixel height space naturally provides an acceleration mechanism for tracing.  
Specifically, SSG~\cite{sheng2022controllable} proposes a ray-scene intersection algorithm in pixel height space. 
Although the ray-scene intersection is designed for tracing visibility, it can be easily modified to trace the closest hit pixel given a ray origin and ray direction in pixel height space. 
In the pixel height space, the ray-scene intersection check happens only along a line between the start and the end pixels. 
The complexity of the ray-scene intersection check in pixel height space is $\mathcal{O}(H)$ or $\mathcal{O}(W)$, without the need to check the intersection with each pixel or each reconstructed triangle. 
Therefore, in practice, we perform ray tracing in the pixel height space in \PHLab. 
We implemented the method using CUDA. It took around 7s to render a noise free reflection for $512 \times 512$ resolution image with 200 samples per pixel. 
Reflection results on different surface materials can be found in Fig.~\ref{fig:reflection}.
Additional examples can be found in \textit{supplementary materials}.

\section{Experiments}\label{Sec:Exp}
Here we show quantitative (the benchmark, metrics for comparison, ablation study) and qualitative evaluation of the buffer channels by comparing to related work.  

\subsection{Quantitative Evaluation of Buffer Channels}
\paragraph{Benchmark:}
To compare our 3D-aware buffer channels fairly with SSN~\cite{sheng2021ssn} that has ground plane assumption, we build two evaluation benchmarks: 1) a ground-shadow benchmark and 2) a wall-shadow benchmark.

The two benchmarks share the same assets, but the ground shadow benchmark only has shadows cast on the ground plane, and the wall shadow benchmark always has part of the shadows cast on walls.
The foreground objects in the assets are composed of 12 new models randomly selected online with different types: robots, animals, humans, bottles, etc. The background objects in the assets are four new backgrounds with different shapes: one wall corner, two wall corners, steps, and curved backgrounds to test the generalization ability to unseen backgrounds. We randomly generate 70 scenes using those unseen foregrounds and background models with different poses of the foreground and background.  

We uniformly sample 125 lights with different positions and different area sizes per scene for the wall shadow benchmark. As shadows on the ground have less variation than the shadows on the wall, we sample eight lights with different positions and different area sizes per scene for the ground shadow benchmark. In total, the ground shadow benchmark is composed of 560 data, and the wall shadow benchmark is composed of 8,750 data. The resolution for each data is $256 \times 256$.

\begin{table}[t]
\centering
\caption{Comparison with SSN~\cite{sheng2021ssn} and SSG~\cite{sheng2022controllable} on the ground-shadow benchmark.}\label{tab:vs_previous}
\small
\begin{tabular}{l|cccc}
\shline
\textbf{Method}          & \textbf{RMSE} $\downarrow$ & \textbf{RMSE-s} $\downarrow$ & \textbf{SSIM} $\uparrow$& \textbf{ZNCC} $\uparrow$\\
\hline
SSN & 0.1207 & 0.1064 & 0.8379 &  0.6118  \\ 
SSG & 0.0254  & 0.0221 & 0.8547 & 0.5679 \\
\textbf{\SN(ours)}  &  \textbf{0.0165}  & \textbf{0.0140} & \textbf{0.9216} & \textbf{0.8180}     \\ 
\shline
\end{tabular}
\label{tab:Quantitative_general}
\end{table}

\begin{figure}[t]
    \centering
    \setlength{\tabcolsep}{1pt}
    \begin{tabular}{cccc}
    \includegraphics[width=0.24\linewidth]{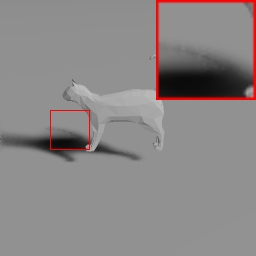}  & 
    \includegraphics[width=0.24\linewidth]{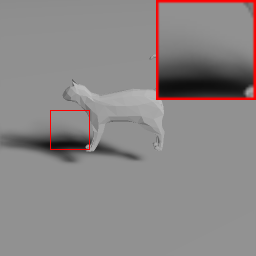} & 
    \includegraphics[width=0.24\linewidth]{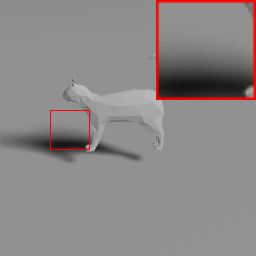} & 
    \includegraphics[width=0.24\linewidth]{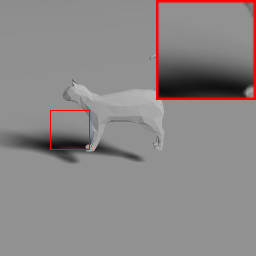} \\
    \includegraphics[width=0.24\linewidth]{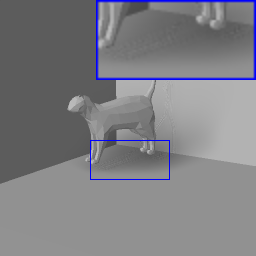} & 
    \includegraphics[width=0.24\linewidth]{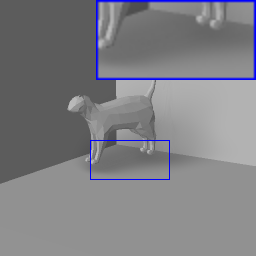} & 
    \includegraphics[width=0.24\linewidth]{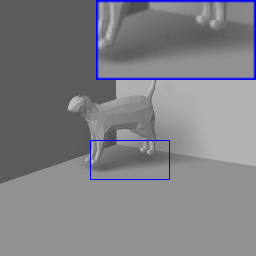} & 
    \includegraphics[width=0.24\linewidth]{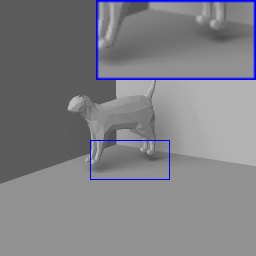} \\
    (a) SSG & (b) SSG-SS & (c) \SN{} & (d) GT 
    \end{tabular}
    \caption{\textbf{Effects of buffer channels.} Best zoom-in. (a) shows the shadows rendered by SSG will be softened uniformly. (b). shows sparse hard shadow channels guide the neural network to be 3D-aware. (c) shows \SN{} can render better quality in the relatively darker regions(note the foot shadow in the second row). (d) is the ground truth shadow rendered by the physically based renderer.}
    \label{fig:ablation}
\end{figure}

\begin{figure*}[t]
\setlength{\tabcolsep}{0.1pt}
\begin{tabular}{cccc}
    \includegraphics[width=0.24\linewidth, trim={1cm 0 2cm 3cm},clip]{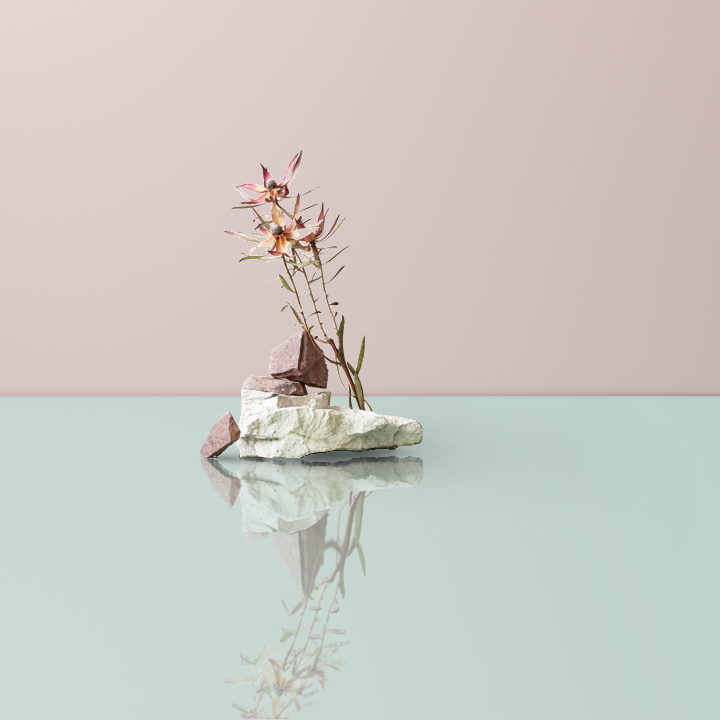} &
    \includegraphics[width=0.24\linewidth, trim={1cm 0 2cm 3cm},clip]{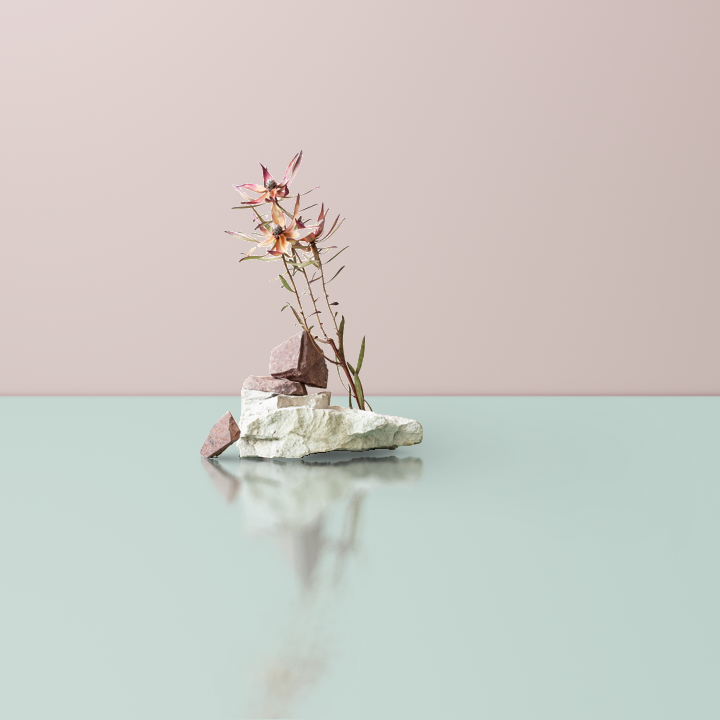} &
    \includegraphics[width=0.24\linewidth, trim={1cm 0 2cm 3cm},clip]{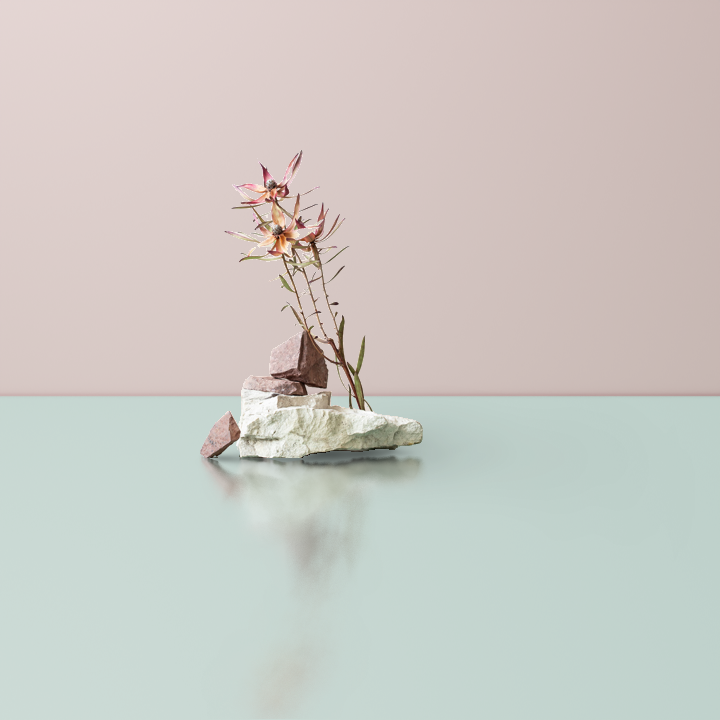} &
    \includegraphics[width=0.24\linewidth, trim={1cm 0 2cm 3cm},clip]{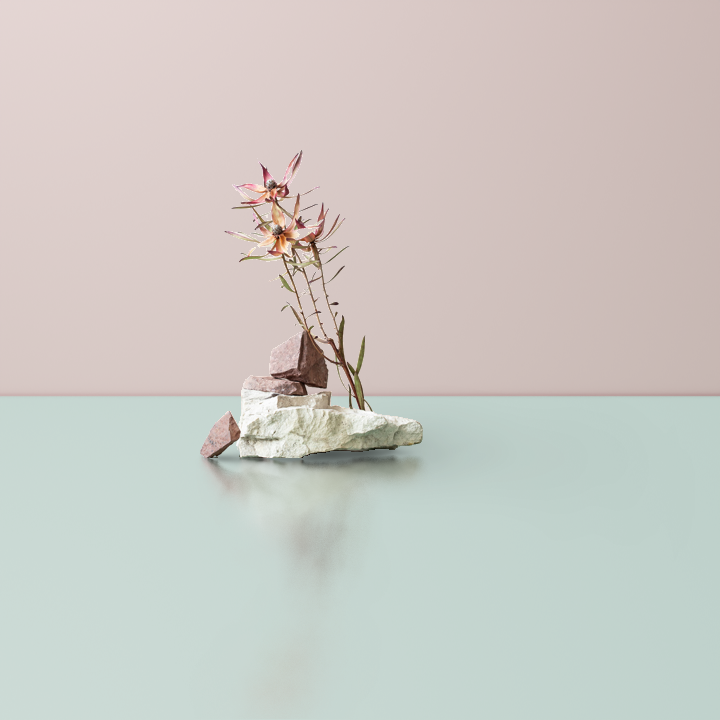} \\ 
    \includegraphics[width=0.24\linewidth, trim={1cm 0 2cm 3cm},clip]{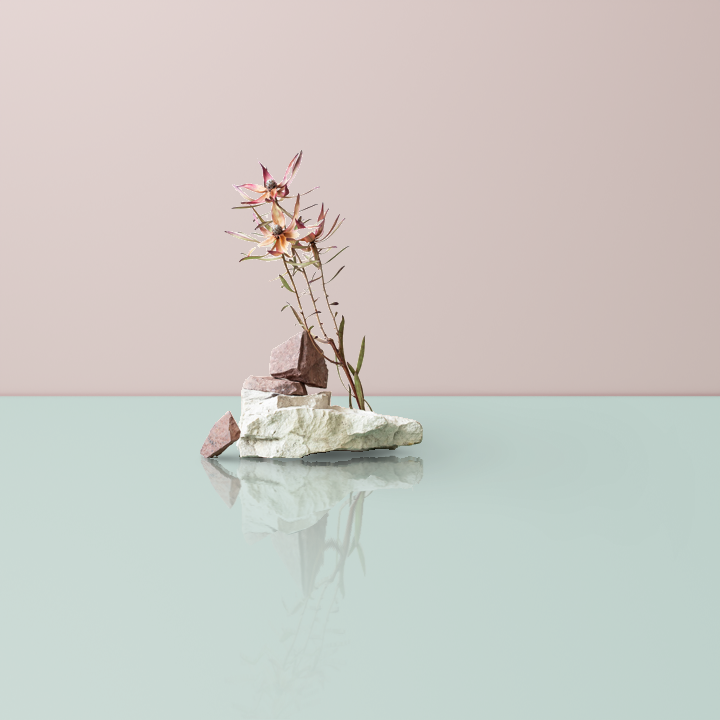} &
    \includegraphics[width=0.24\linewidth, trim={1cm 0 2cm 3cm},clip]{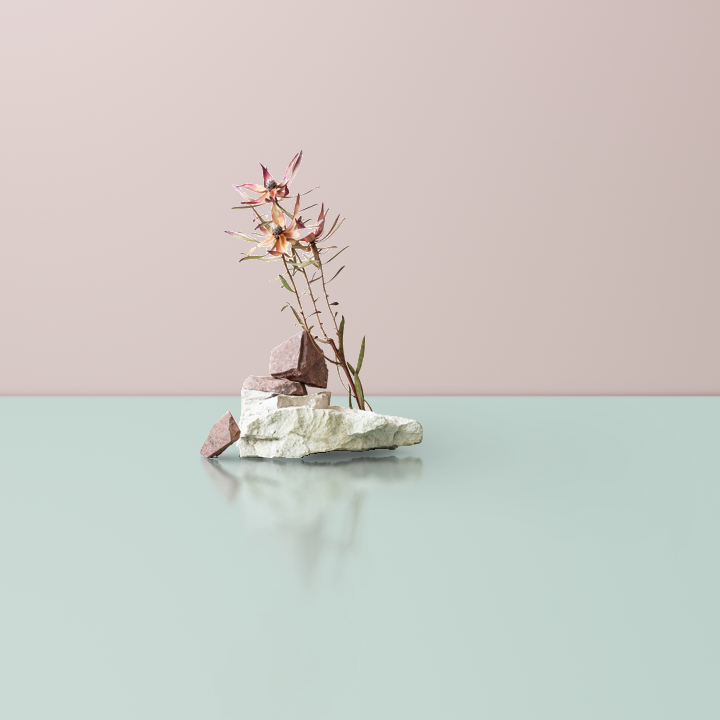} &
    \includegraphics[width=0.24\linewidth, trim={1cm 0 2cm 3cm},clip]{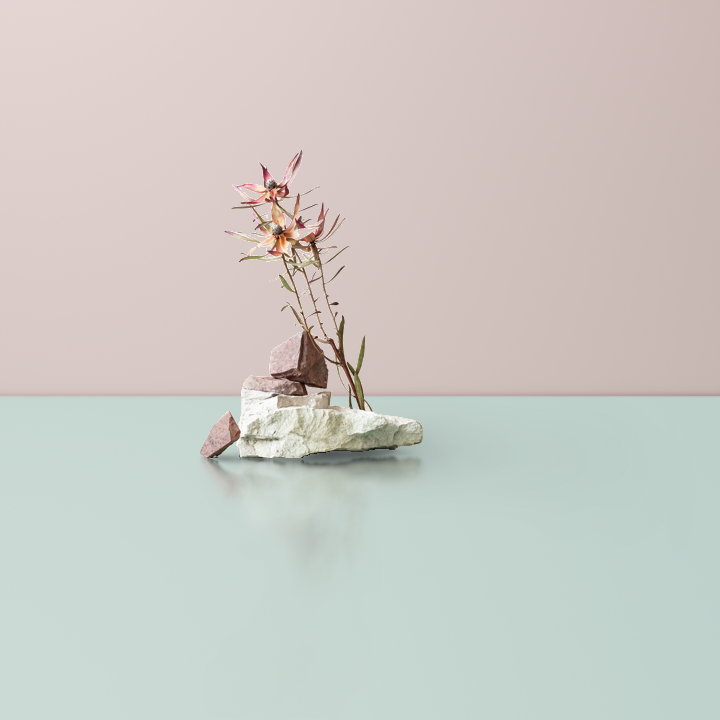} &
    \includegraphics[width=0.24\linewidth, trim={1cm 0 2cm 3cm},clip]{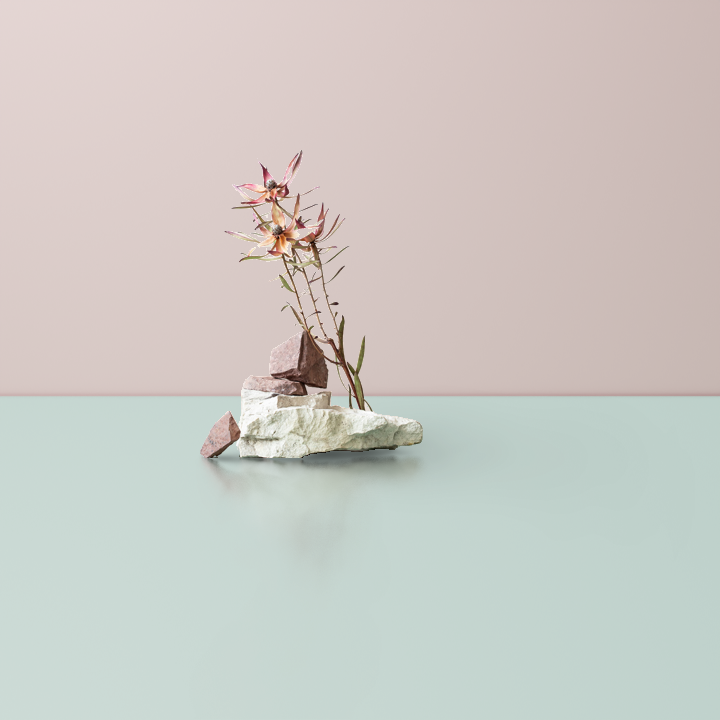} \\ 
 \end{tabular}
 \caption{\textbf{Reflection.} \PHLab{} can render reflection with different physical materials. From left to right, the ground surface glossness increases. The top to bottom, the ground uses different $\eta$ in Fresnel effects.}
 \label{fig:reflection}
 \vspace{-1em}
\end{figure*}

\begin{figure}[t]
    \centering
    \includegraphics[width=0.49\linewidth]{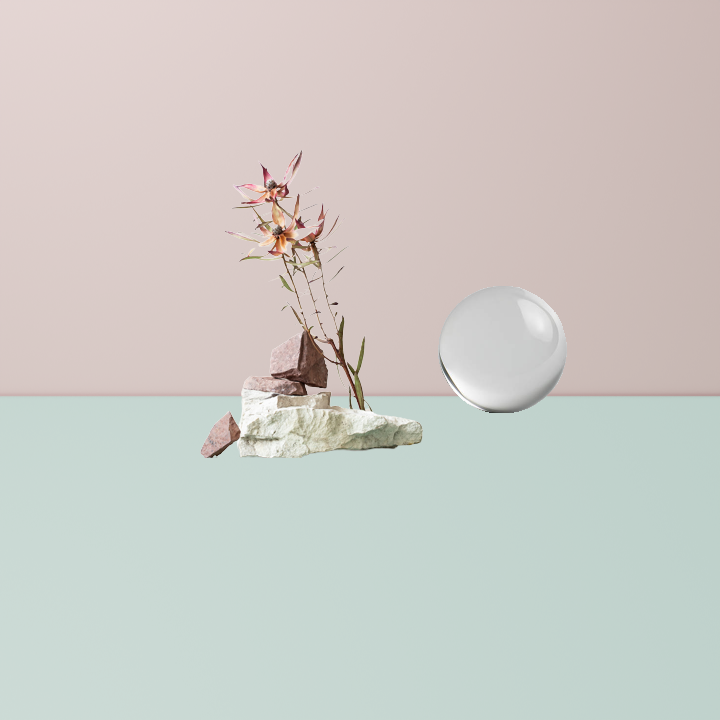}
    \includegraphics[width=0.49\linewidth]{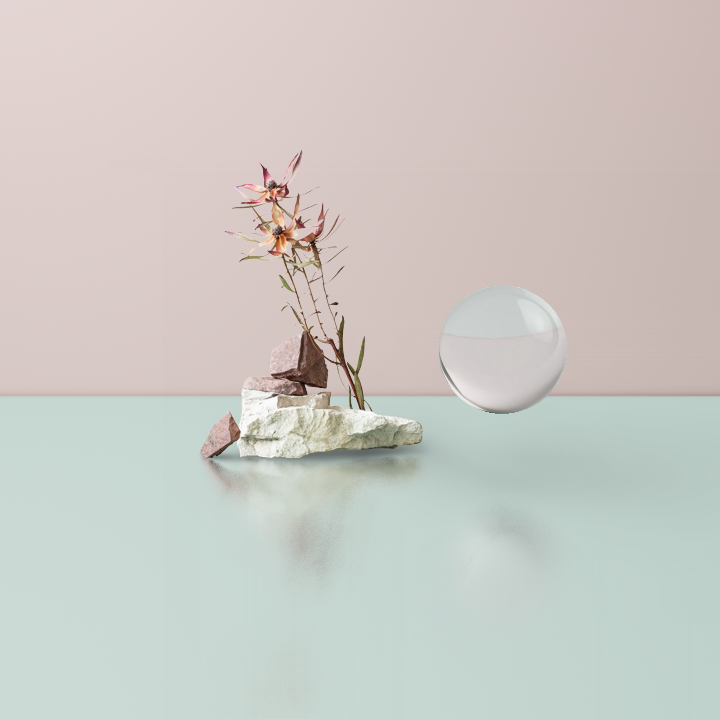}
    \caption{\textbf{Refraction.} Given the cutout and the background in the left image, the refraction lighting effect for the crystal ball can also be rendered by \PHLab{}.}
    \label{fig:refraction}
    \vspace{-1em}
\end{figure}

\paragraph{Metrics:} We use the per-pixel metric RMSE and a scale-invariant RMSE-s~\cite{sun_single_2019}. Similar to~\cite{sun_single_2019}, shadow intensity may vary, but the overall shapes are correct. We also used perception-based metrics SSIM and ZNCC to measure shadow quality. Our SSIM implementation uses $11 \times 11$ Gaussian filter with $\sigma=1.5, k_1=0.01, k_2=0.03$. 

\begin{table}[t]
\centering
\caption{Result on the wall-shadow benchmark. We show the effectiveness of each buffer channel. SSG-BH: SSG with background pixel height. SSG-D: SSG with XYH distance channel. SSG-D-BH: SSG with XYH distance and background pixel height. SSG-SS: SSG with the sparse shadow channel. SSG-SS-BH: SSG with sparse shadow channel and background pixel height. \SN: SSG with all the buffer channels.}\label{tab:ablation_study}
\small
\begin{tabular}{l|cccc}
\shline
\textbf{Method}          & \textbf{RMSE} $\downarrow$ & \textbf{RMSE-s} $\downarrow$ & \textbf{SSIM} $\uparrow$& \textbf{ZNCC} $\uparrow$\\
\hline
SSG        & 0.0242  & 0.0209 & 0.8561 & 0.6460 \\
\hline
SSG-BH     & 0.0248  & 0.0207 & 0.8587 & 0.6506 \\
SSG-D      & 0.0230  & 0.0210 & 0.8739 & 0.6499  \\
SSG-D-BH   & 0.0231  & 0.0201 & 0.8752 & 0.6719   \\
SSG-SS     & 0.0164  & 0.0149 & 0.9139 & 0.8228   \\
SSG-SS-BH  & 0.0184  & 0.0158 & 0.9136 & 0.8029  \\
SSG-SS-D   & 0.0159  & 0.0139 & 0.9153 & 0.8494  \\
\textbf{SSN++(ours)} & \textbf{0.0153} & \textbf{0.0136} & \textbf{0.9277} & \textbf{0.8575}\\
\shline
\end{tabular}
\end{table}

\begin{figure}[t]
    \centering
    \includegraphics[width=0.48\linewidth, trim={2cm 1cm 10cm 6cm}, clip]{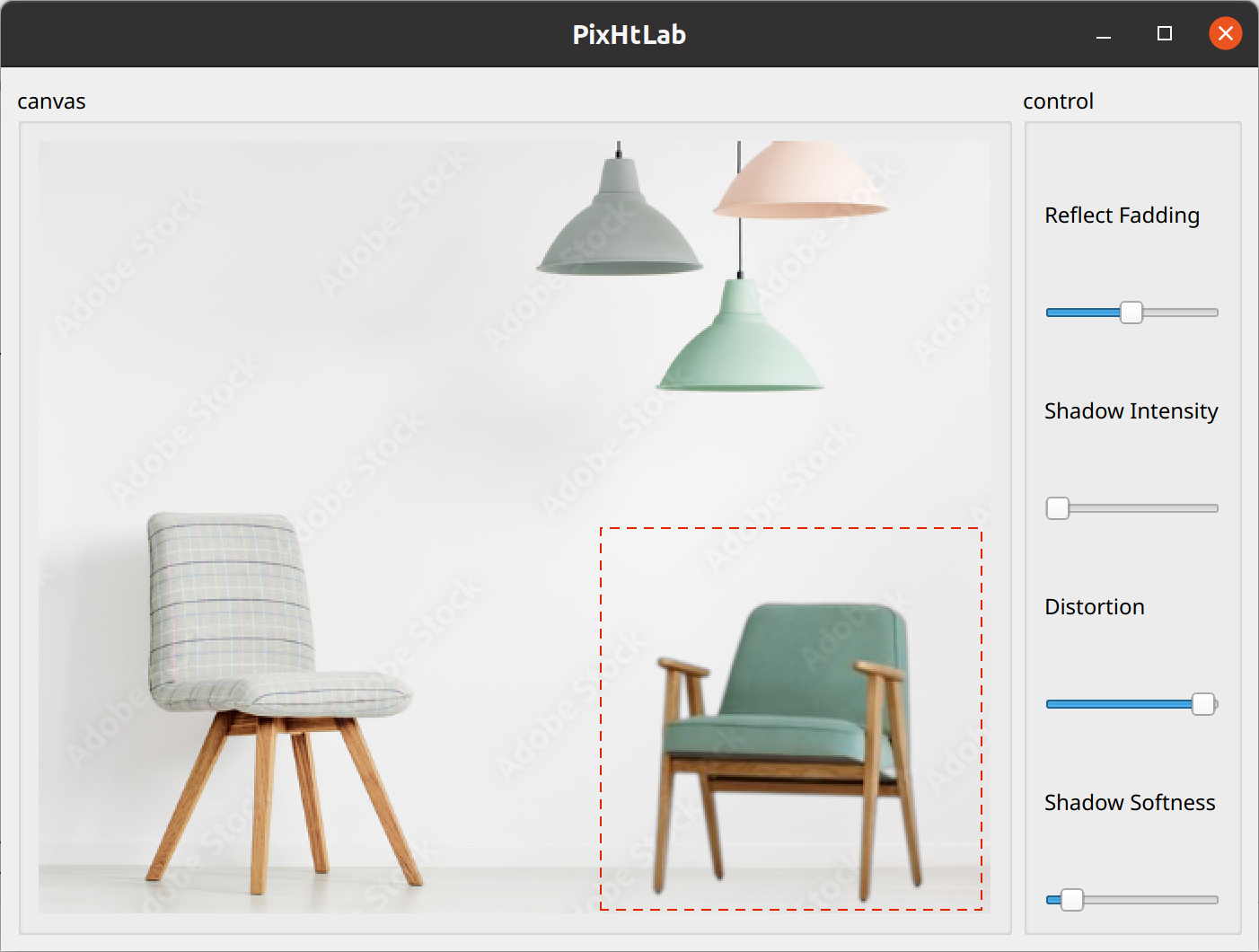}
    \includegraphics[width=0.48\linewidth, trim={2cm 1cm 10cm 6cm}, clip]{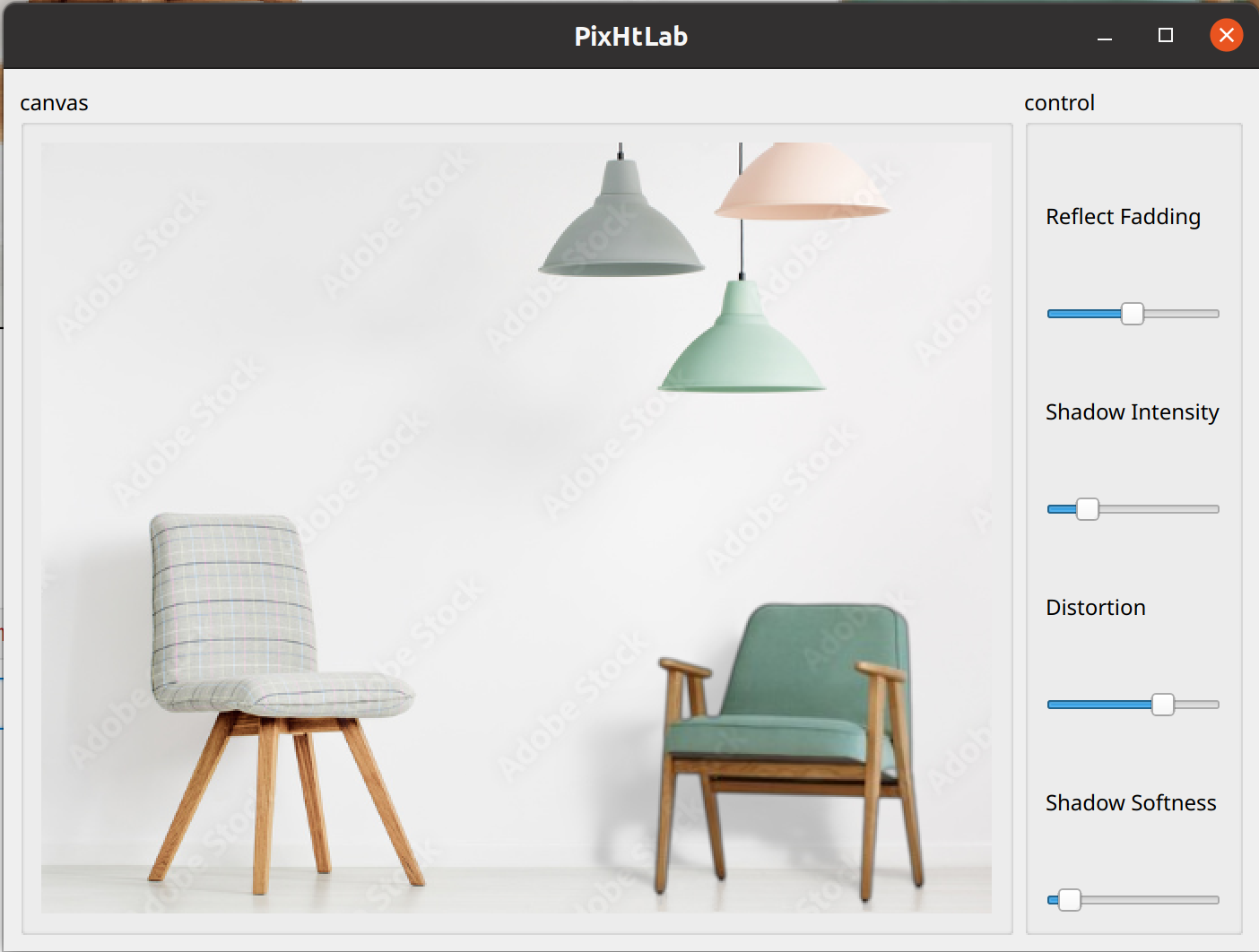}
    \includegraphics[width=0.48\linewidth, trim={2cm 1cm 10cm 6cm}, clip]{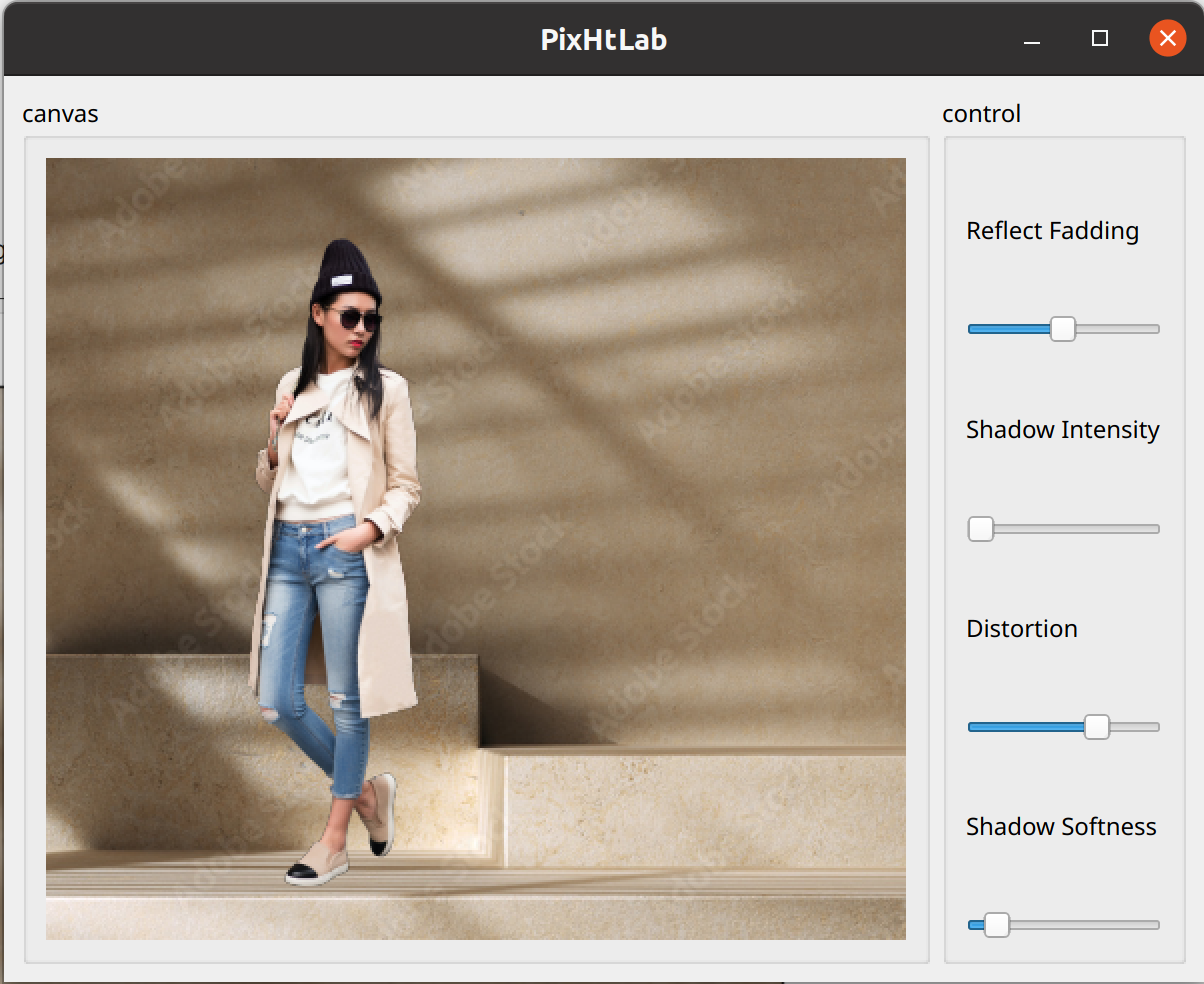}
    \includegraphics[width=0.48\linewidth, trim={2cm 1cm 10cm 6cm}, clip]{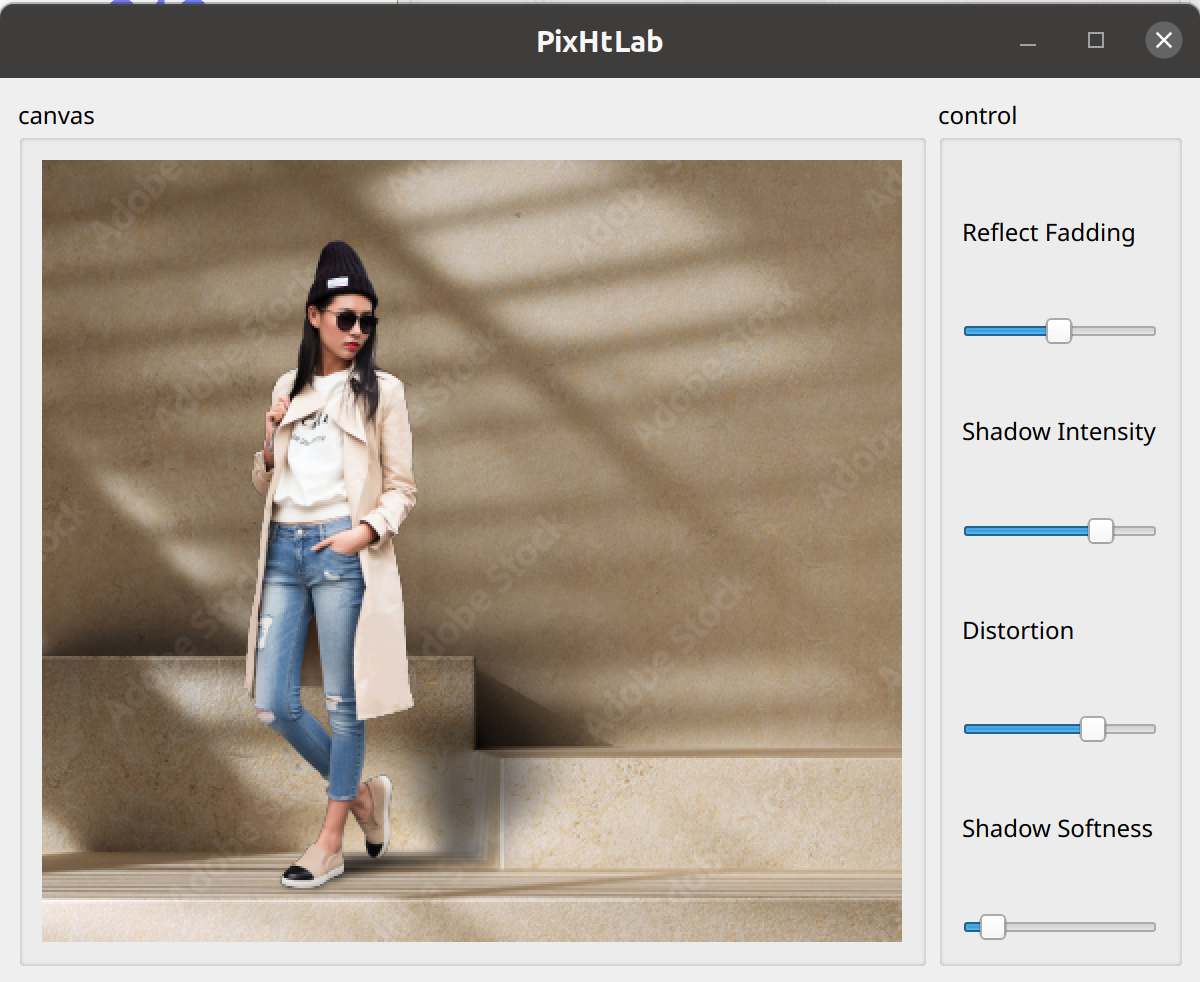}
    \caption{Real foreground and background examples created with our GUI. Zoom in for the best view. \textit{Credit: Adobe Stock.}}
    \label{fig:real-backgrounds} \vspace{-3mm}
\end{figure}

\begin{figure*}[t]
\setlength{\tabcolsep}{0.0pt}
\renewcommand{\arraystretch}{0.0}
\begin{tabular}{cccc}
    \includegraphics[width=0.24\linewidth, trim={1cm 2.5cm 3cm 1.5cm},clip]{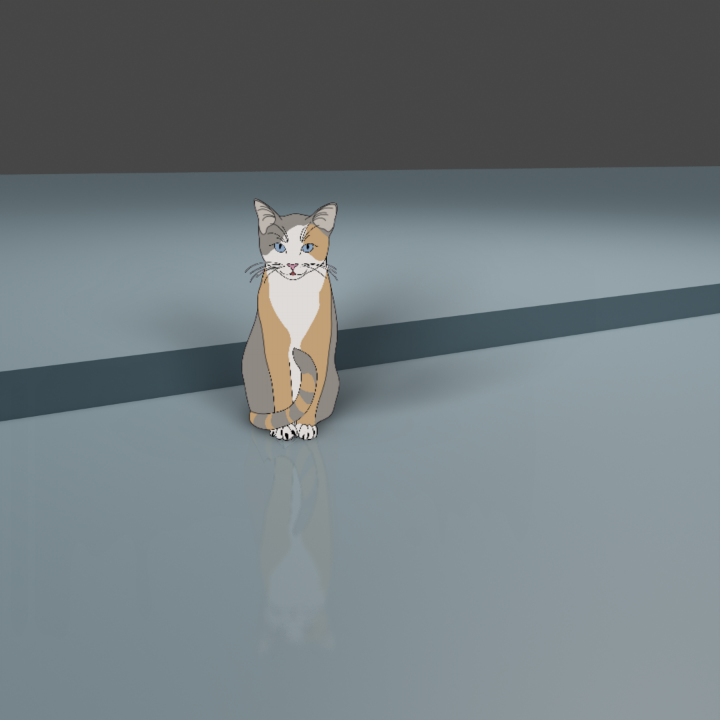} &
    \includegraphics[width=0.24\linewidth, trim={2.0cm 0.2cm 2.0cm 3.8cm},clip]{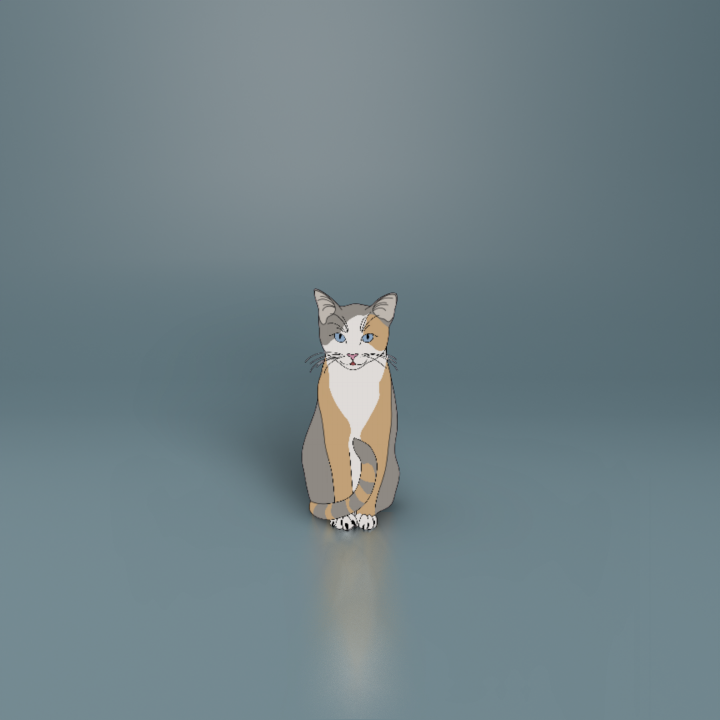} &
    \includegraphics[width=0.24\linewidth, trim={0.4cm 1.2cm 2.4cm 1.6cm},clip]{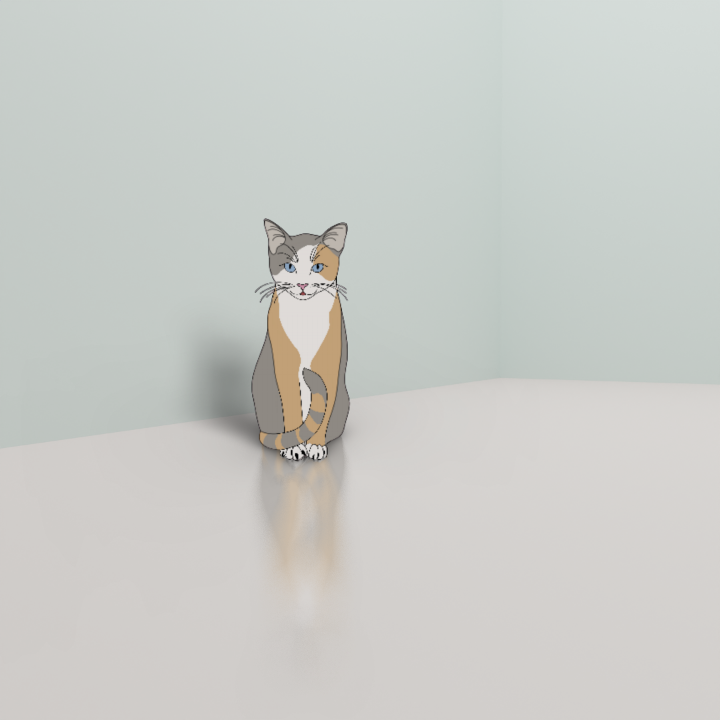} & 
    \includegraphics[width=0.24\linewidth, trim={1.3cm 2.6cm 3.3cm 2.0cm},clip]{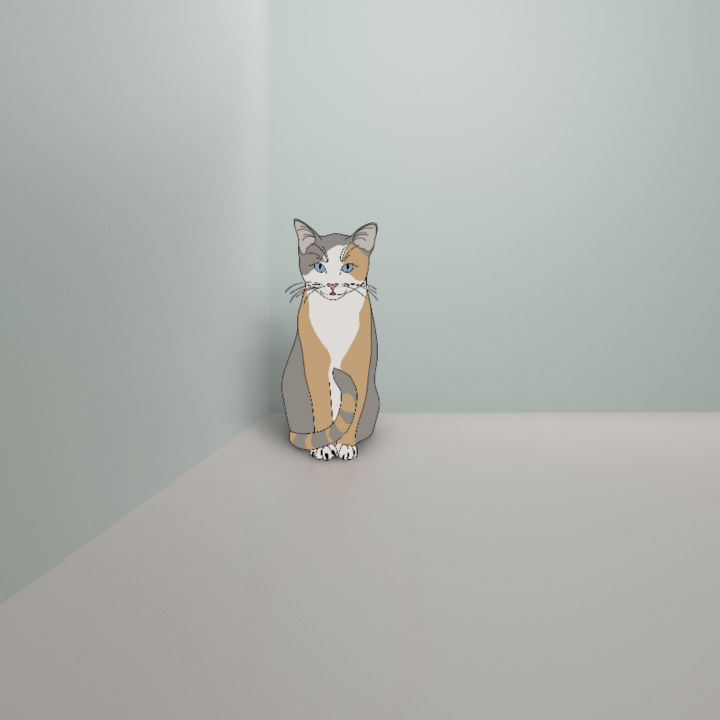} \\ 
    \includegraphics[width=0.24\linewidth, trim={1cm 2cm 3cm 2cm},clip]{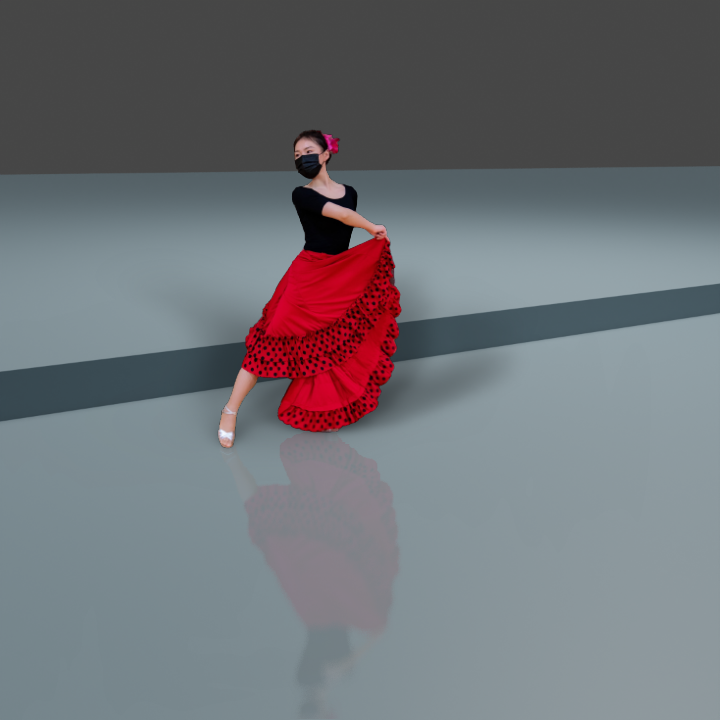} &
    \includegraphics[width=0.24\linewidth, trim={2.3cm 0.1cm 2.3cm 4.5cm},clip]{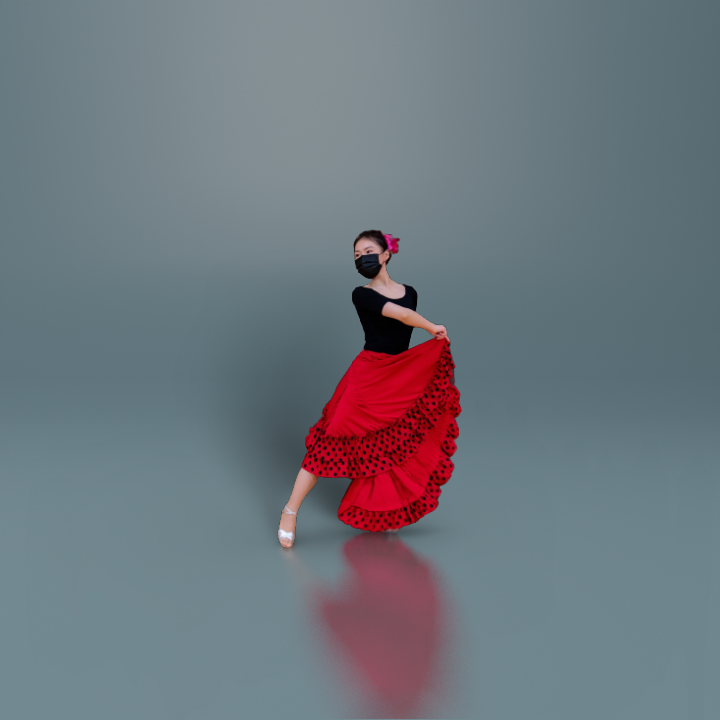} &
    \includegraphics[width=0.24\linewidth, trim={1cm 2.5cm 4cm 2.5cm},clip]{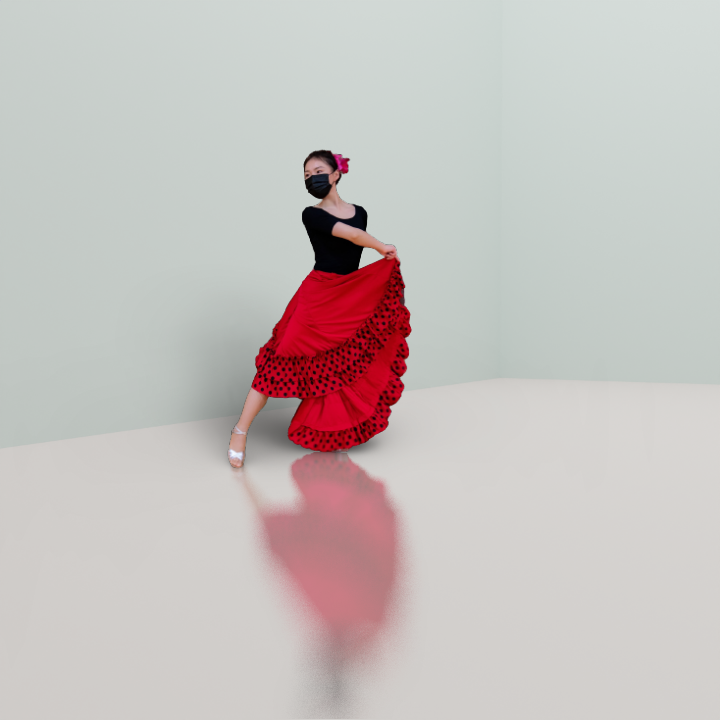} & 
    \includegraphics[width=0.24\linewidth, trim={1cm 2.5cm 4cm 2.5cm},clip]{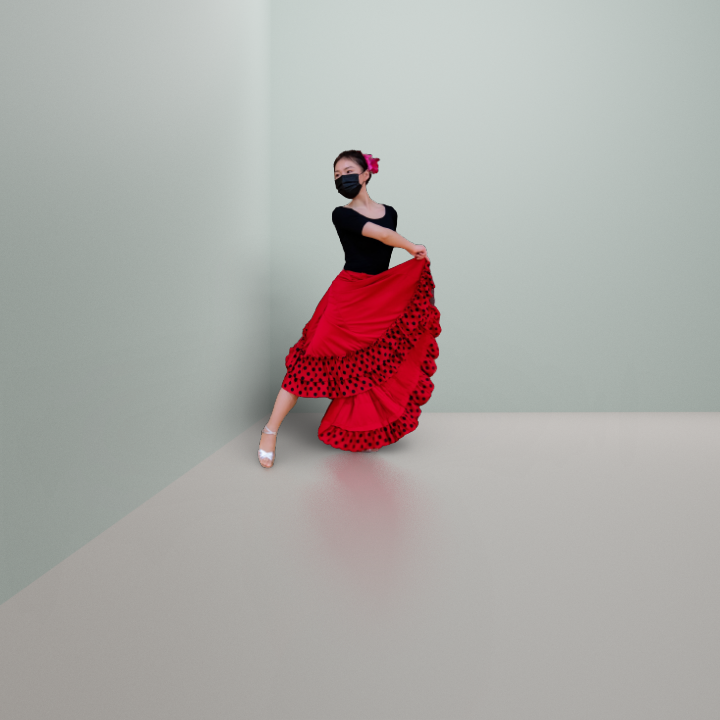}  \\ 
    \includegraphics[width=0.24\linewidth, trim={1.1cm 3cm 3.1cm 1.2cm},clip]{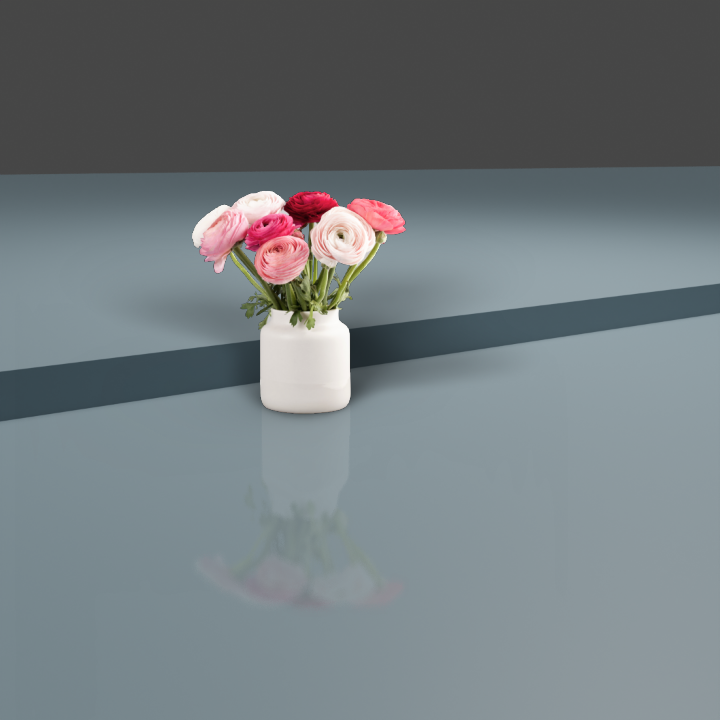} &
    \includegraphics[width=0.24\linewidth, trim={2cm 0cm 2cm 4cm},clip]{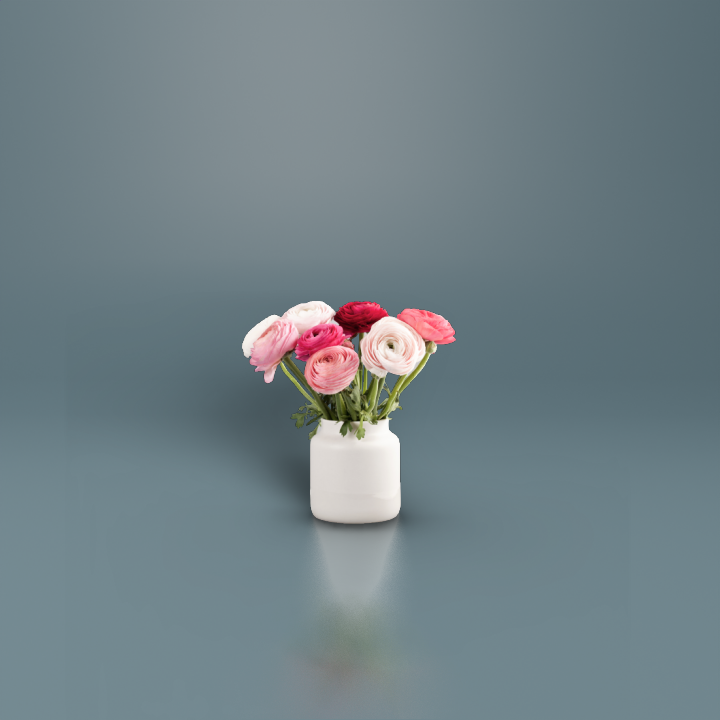} &
    \includegraphics[width=0.24\linewidth, trim={0.9cm 2.4cm 3.9cm 2.4cm},clip]{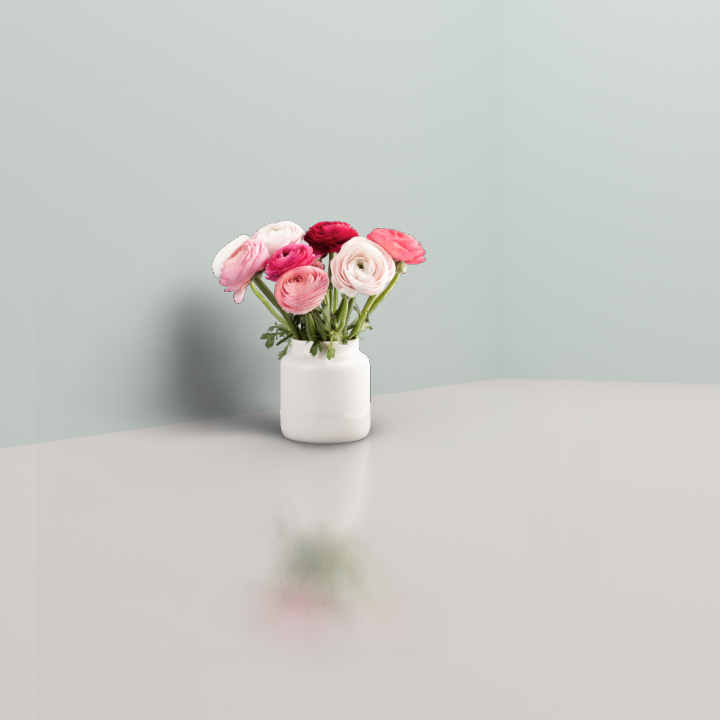} & 
    \includegraphics[width=0.24\linewidth, trim={1.6cm 1.6cm 3.2cm 3.2cm},clip]{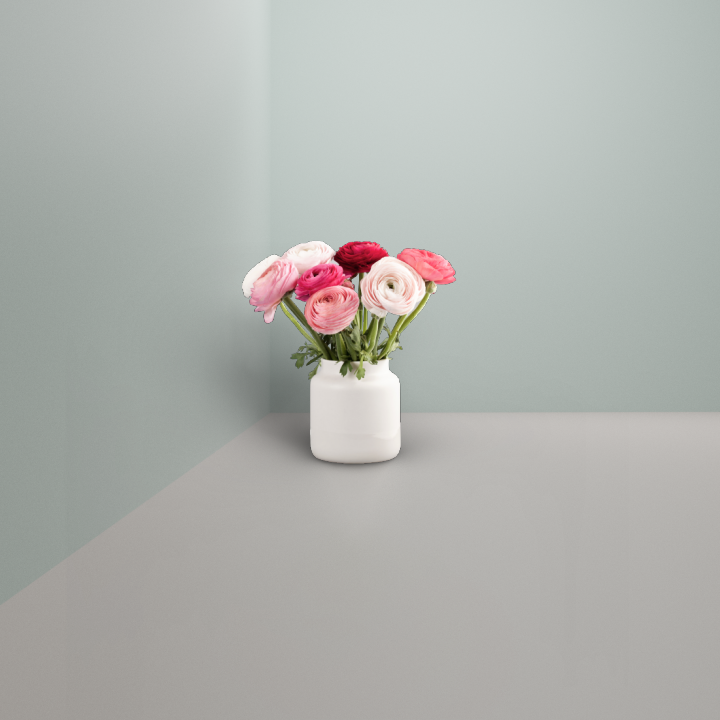} \\ 
 \end{tabular}
 \caption{\textbf{More results.} \PHLab{} is agnostic to the cutout object categories. Lighting effects can be generated to general backgrounds. \PHLab{} can also generate multiple soft shadows shown in the first column. The first column uses a step background. The second column uses a curved background. The third column uses a L shape wall background. The forth column uses a corner background. }
 \label{fig:more results}
\end{figure*}

\paragraph{\SN{} on the ground-shadow benchmark.}
As SSN has a ground plane assumption, we use the ground-shadow benchmark to compare fairly. We compare our \SN{} with SSN and the other soft shadow rendering network SSG proposed recently on the ground-shadow benchmark. Results are shown in Tab.~\ref{tab:vs_previous}. Our \SN{} outperforms SSN and SSG in all metrics. Compared with the STOA SSG, \SN{} improves RMSE by $35\%$, RMSE-s by $36\%$, SSIM by $7.8\%$, ZNCC by $44\%$. The statistics show in the simplest ground plane shadow receiver case, \SN{} still has significant improvement. Even in the simplest case, buffer channels significantly improve the soft shadow rendering quality. 

\paragraph{\SN{} on the wall-shadow benchmark.}
\SN{} and SSG share the same backbone, but they are guided by different buffers. Therefore, we treat SSG as the basic baseline and do the ablation study together in this section. Results are shown in Tab.~\ref{tab:ablation_study}.

Our proposed \SN{} outperforms all the other methods guided by other subsets of the buffer channels. Each buffer channel outperforms SSG in all the metrics, showing that those 3D-aware buffer channels are useful to guide the SSG to render better soft shadows.
SSG-D-BN fixes more errors than SSG-D or SSG-BN, showing that the combination of relative distance in pixel height space helps the neural network improve the soft shadow quality. 
SSG-SS significantly outperforms all the previous baselines by improving RMSE by $29\%$ and SSIM by $4\%$ than SSG-D-BH, which shows that the sparse shadows channel plays the most important role in guiding the SSG to render soft shadows. 
Combining the sparse shadow channel with the relative distance channel only improves RMSE by $3\%$ and SSIM by $0.15\%$ than only using the sparse shadow channel as additional channels while combing the sparse shadow channel with the background pixel height channel performs worse than only using the sparse shadow channels as an additional channel for SSG, with RMSE degraded by $12\%$ and SSIM by $0.03\%$. 

Our \SN{} combines the sparse shadow channel, the relative distance channel, and the background pixel height channel together and achieves the best performance, improving in all the metrics significantly. 
Compared with SSG, our \SN{} improves RMSE by $38\%$, RMSE-s by $33\%$, SSIM by $8\%$ and ZNCC by $32\%$. 

\subsection{Qualitative Evaluation of Buffer Channels}
\paragraph{Effects of buffer channels.}
Fig.~\ref{fig:ablation} shows the effects of the buffer channels. 
Fig.~\ref{fig:ablation} (b). shows the sparse shadow guides the neural network to render better contour shapes as the sparse hard shadows are samples from the outer contour of the shadow regions. 
However, when the geometry has complex shapes, and the sparse hard shadows are mixed together, e.g., the foot regions of the cat in the second row of Fig.~\ref{fig:ablation}, the relative spatial information is ambiguous. 
The relative distance map can further guide \SN{} to keep the regions close to the objects dark instead of over soft(See Fig.~\ref{fig:ablation} (c) in the second row.).

\section{Discussion}
\paragraph{Light effects generated by \PHLab{}.}
\PHLab{} can reconstruct the surface normal solely based the pixel height inputs. As discussed in 

\PHLab{} does not have assumptions on the cutout object types and background types. 
No matter for realistic cutouts or cartoon cutouts, studio background or real world background, \PHLab{} can render the light effects. 
(see Fig.~\ref{fig:real-backgrounds} and Fig.~\ref{fig:more results} and the demo video showing the \PHLab{} system in the \textit{supplementary materials}). Similar to SSG, \PHLab{} allows the user to intuitively control the shadow direction and softness, control the horizon position to tweak the shadow distortion, and change parameters to control the physical parameters of the reflection. Our methods can also be applied to multiple object compositing and multiple shadows. Please refer to \textit{supplementary materials} for more examples. 

There exist more potential additions for \PHLab{} and other light effects such as refraction(see Fig.~\ref{fig:refraction}) could be implemented. Parameters related to the refraction surface, like the refraction index, can be controlled as well. We demonstrate more results in the supplementary material.  %

\paragraph{Limitation.}
As \PHLab{} is based on the pixel height map and the common limitations for the pixel height representation apply to our methods as well. One of them is that it takes the image cutout as the proxy of the object and the back face or hidden surface contributing to the light effect generation is missing. A back face prediction neural network can be explored to address this problem. 
Another limitation specific to \PHLab{} is that the proposed method uses the cutout color as the reflected color, which is not precise for cases when the surface has view-dependent colors.

\section{Conclusion and Future Work}
We propose a novel system \PHLab{} for generating perceptually plausible light effects based on the pixel height representation. 
The mapping between the 2.5D pixel height and 3D has been presented to reconstruct the surface geometry directly from the pixel height representation.
Based on the reconstruction, more light effects, including physically based reflection and refraction, can be synthesized for image compositing.  
Also, a novel method \SN{} guided by 3D-aware buffer channels is proposed to improve the soft shadow quality that is cast on general shadow receivers. 
Quantitative and qualitative experiments demonstrate that the results and generalization ability of the proposed \SN{} significantly outperforms previous deep learning-based shadow synthesis methods. 
However, our \PHLab{} synthesize the light effect solely based on the cutout colors. A back face prediction neural network may address the issue and is worth future exploration.  

{\small
\bibliographystyle{ieee_fullname}
\bibliography{egbib}
}

\end{document}